%% file: sample-sigplan.tex
\documentclass[sigplan,nonacm]{acmart}

\AtBeginDocument{%
  }

\usepackage{graphicx}
\usepackage{forest}
\usepackage{subcaption}
\usepackage{textcomp}
\usepackage{tikz}
\usepackage{xcolor}
\usepackage{multirow}
\usepackage{algorithm}
\usepackage{algpseudocode}
\usepackage{pgfplots}
\usepackage{lipsum} % For dummy text
\usepackage{booktabs} % For professional looking tables (toprule, midrule, bottomrule)
\usepackage{siunitx}  % For aligning numbers by decimal point and S column type
\usepackage{rotating} % For sidewaystable if needed
\usepackage{tabularx}
\usepackage{multirow}
\usepackage{listings}
\usepackage{wrapfig}
\usepackage{colortbl}
\usepackage{tcolorbox}
\tcbuselibrary{listings}

\setcopyright{acmlicensed}
\copyrightyear{2018}
\acmYear{2018}
\acmDOI{XXXXXXX.XXXXXXX}
\acmConference[Conference acronym 'XX]{Make sure to enter the correct
  conference title from your rights confirmation email}{June 03--05,
  2018}{Woodstock, NY}
\acmISBN{978-1-4503-XXXX-X/2018/06}

\begin{document}

\title{Toward Compiler World Models: Learning Latent Dynamics for Efficient Tensor Program Search}

\author{
Haolin Pan\textsuperscript{1,2,3},
Lianghong Huang\textsuperscript{2,3},
Xulin Zhou\textsuperscript{2,3},
Mingjie Xing\textsuperscript{1,2,3,*},
Yanjun Wu\textsuperscript{2,3} \\
\textsuperscript{1}\textit{Hangzhou Institute for Advanced Study, University of Chinese Academy of Sciences, Hangzhou, China} \\
\textsuperscript{2}\textit{Institute of Software, Chinese Academy of Sciences, Beijing, China} \\
\textsuperscript{3}\textit{University of Chinese Academy of Sciences, Beijing, China} \\
\{panhaolin21,huanglianghong25\}@mails.ucas.ac.cn,\\
\{zhouxulin2023, mingjie, yanjun\}@iscas.ac.cn
}

% \maketitle

\renewcommand{\shortauthors}{Haoin Pan et al.}

\begin{abstract}
Tensor program optimization is critical for modern machine learning systems, but finding the most efficient program is highly challenging due to the massive search space. To avoid expensive on-device measurements, existing auto-schedulers use learned cost models to evaluate candidate programs. However, these models typically treat each candidate as a static code snapshot. This static view ignores the step-by-step transformation process that actually created the program. Consequently, the evaluator cannot understand how a scheduling decision made in a later step depends on the context of earlier ones. Furthermore, it is easily misled by superficial code variations, failing to recognize when two structurally different programs actually perform identically on hardware.

To bridge this gap, we introduce a \emph{world-model-inspired perspective} that formulates schedule evaluation as action-conditioned latent dynamics over program states. Recognizing that compiler optimization is essentially a sequence of structured algebraic transformations, we design a learning-based evaluation framework that simulates these state transitions in a continuous latent space. Starting from an initial program state, our framework rolls out the scheduling actions step-by-step using a lightweight transition model, predicting the terminal-state representation without invoking slow AST mutations or text-based encoding. This dynamic representation is then combined with structured action and hardware features to rank candidates. We instantiate this framework in TVM AutoScheduler using a learned state encoder, an action-conditioned transition model, and a ranking cost model.

On an Intel Xeon Gold 6430 CPU and an NVIDIA RTX 4090 GPU, our method improves representative-subgraph latency over Ansor by 1.37$\times$ on GPU and 1.54$\times$ on CPU under the same 64-trial budget. It slightly outperforms Ansor-10K in arithmetic mean and matches it within 2.2\% in geometric mean using 10$\times$ fewer measurements, and speeds up full-model inference over PyTorch/PyTorch-opt(cuDNN) by 4.61$\times$/3.67$\times$ geometric mean and up to 58.61$\times$/35.50$\times$.
\end{abstract}

\maketitle
\textsuperscript{*}Corresponding author.


\keywords{Do, Not, Use, This, Code, Put, the, Correct, Terms, for,
  Your, Paper}

\section{Introduction}

Efficient tensor program optimization is an important component of modern machine learning systems. To obtain high performance across different workloads and hardware targets, deep learning systems rely on either vendor-provided kernel libraries, such as cuDNN \cite{chetlur2014cudnn} and oneDNN \cite{onednn-docs}, or tensor compilers that generate optimized tensor programs automatically. Because manually engineered kernel libraries are costly to build and maintain, tensor compilers have become an important approach for producing optimized implementations across evolving models and hardware platforms.

A tensor compiler typically includes a graph-level frontend, a tensor-program optimization framework, and a backend code generator. The frontend converts models from frameworks such as TensorFlow \cite{abadi2016tensorflow}, PyTorch \cite{ansel2024pytorch} into a unified intermediate representation, applies graph-level optimizations, and partitions the workload into subgraphs. The tensor-program optimization framework then takes these subgraphs as inputs and makes scheduling decisions such as loop transformations, tiling, memory placement, parallelization, vectorization, and hardware binding to produce tensor programs. Finally, the backend lowers these programs into target-specific executables. Although this workflow reduces the need for manual kernel engineering, the optimization problem remains difficult because the search space is combinatorial, direct performance measurement is expensive, and ineffective guidance can consume substantial tuning budget.

Prior work has improved tensor program tuning mainly from two directions. One direction focuses on exploration. Classical auto-schedulers, template- and heuristic-based systems, and large-space frameworks such as AutoTVM \cite{autotvm}, Ansor \cite{zheng2020ansor}, MetaSchedule \cite{MetaSchedule}, and AKG \cite{zhao2021akg} improve how candidate schedules are generated and searched. Another direction focuses on evaluation. Learned cost models such as TenSet \cite{zheng2021tenset} and TLP \cite{zhai2023tlp} improve candidate ranking before expensive on-device measurement. TLM \cite{TLM} also improves exploration by generating scheduling decisions over large decision spaces. These studies have advanced tensor program search, but candidate schedules are still commonly evaluated as \emph{static snapshots of code}. This static view ignores the step-by-step transformation process that created the program. Consequently, existing evaluators struggle to capture how subsequent optimization decisions depend on the context of earlier ones, and they are easily misled by superficial code variations in structurally equivalent programs.

To address these limitations, we introduce a \emph{world-model-inspired perspective} for tensor program evaluation. A world model is a predictive surrogate designed to learn the transition dynamics of an environment. It maps complex, high-dimensional, and often noisy observations into a structured continuous latent space \cite{lecun2022path,bi2026motus,taniguchi2026generative}, where superficial variations are suppressed and semantically meaningful states are represented as distinct geometric coordinates. Inspired by this perspective, we model tensor program optimization as a latent state-transition process: the compiler and target hardware form the environment, TensorIR program representations correspond to states, and scheduling transformations act as transition operators that evolve those states over time.
Instead of evaluating a candidate based only on its final, static text, our framework models the state transitions in a continuous representation space. Starting from the initial unoptimized program, the model simulates the cumulative effects of each scheduling decision step-by-step. This latent-space rollout allows the evaluator to predict the final program's quality efficiently, bypassing the need to explicitly materialize and encode every intermediate TensorIR state during candidate evaluation. Since the primary role of the evaluator is to prioritize candidates within the same workload before expensive physical measurement, the final stage is formulated as workload-level candidate ranking. We implement this formulation in TVM AutoScheduler, where it is instantiated by a TensorIR encoder, a multi-step latent state predictor, and a ranking cost model.

We evaluate the proposed framework on an Intel Xeon Gold 6430 CPU and an
  NVIDIA GeForce RTX 4090 GPU using seven neural-network models and 22
  representative tensor-program subgraphs.  Under the same 64-trial
  budget, our method outperforms Ansor on 21 of 22 GPU representative
  subgraphs and all 22 CPU representative subgraphs, reaching up to
  2.40$\times$ and 3.76$\times$ speedup, respectively.  On weighted
  model-level latency, our method achieves 1.22$\times$ geometric-mean
  speedup on GPU and 1.62$\times$ on CPU, with per-model speedups up to
  1.55$\times$ and 1.88$\times$.  More importantly, our 1K-trial search
  slightly outperforms Ansor-10K in arithmetic mean and matches it within
  2.2\% geometric-mean latency on representative subgraphs, achieving
  comparable search quality with 10$\times$ fewer measurements.
  Full-model results show 4.61$\times$/3.67$\times$ geomean and
  58.61$\times$/35.50$\times$ peak speedups over
  PyTorch/PyTorch-opt(cuDNN).

In summary, this paper makes the following contributions:
\begin{itemize}

\item We adapt the world-model perspective to tensor optimization and establish a compiler-world-model framework for tensor program evaluation, where candidate schedules are modeled through action-conditioned latent state evolution.

\item We construct a TVM/TenSet-based state-prediction dataset for learning compiler state transitions. Built from tuning logs and aligned TensorIR states, the dataset organizes pre-schedule states, scheduling-action sequences, intermediate states, and post-schedule states into action-state trajectories for multi-step program-state prediction.

\item We implement the proposed framework in TVM AutoScheduler and demonstrate that it substantially reduces autotuning cost. Across diverse workloads on CPU and GPU platforms, our method achieves search quality comparable to Ansor using 10$\times$ fewer measurements and delivers significant end-to-end performance improvements.
\end{itemize}

\section{Background and Motivation}

\subsection{Background}

\subsubsection{Tensor-Program Optimization Pipeline}

\begin{figure}[t]
    \centering
    \includegraphics[width=0.91\linewidth]{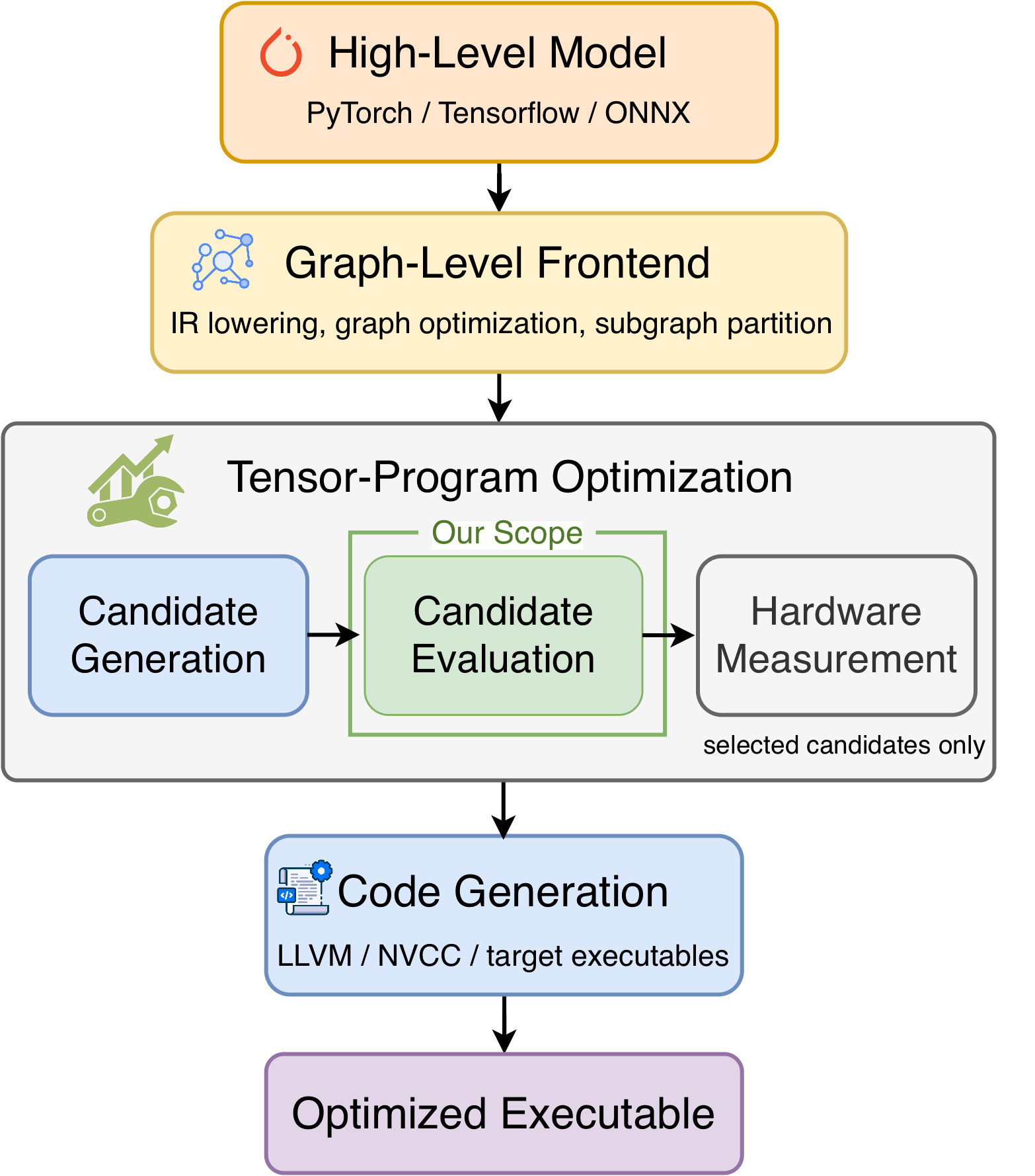}
    \caption{Common deep learning compiler pipeline. Our work focuses on candidate evaluation.}
    \label{fig:tensor-compiler}
\end{figure}

Common deep learning compilers translate high-level models into hardware-efficient low-level implementations through a multi-stage compilation pipeline \cite{chen2018tvm,Halide,Tensorcomprehensions,zheng2020flextensor,NeonCPU}. As illustrated in Figure~\ref{fig:tensor-compiler}, this pipeline typically includes a graph-level frontend, a tensor-program optimization framework, and a backend code generator. The frontend converts models from frameworks (e.g., TensorFlow \cite{abadi2016tensorflow}, PyTorch \cite{ansel2024pytorch}, MXNet \cite{chen2015mxnet}) into a compiler intermediate representation \cite{onnx-docs}, applies graph-level optimizations, and partitions the workload into subgraphs \cite{roesch2018relay}. The backend then lowers optimized tensor programs into target-specific executables (e.g., LLVM \cite{llvm-docs}, NVCC \cite{nvcc-docs}).

Between these two stages lies tensor-program optimization, which is responsible for generating efficient implementations for individual subgraphs. At this stage, the compiler must make a sequence of scheduling decisions, such as loop tiling, reordering, fusion, memory placement, parallelization, vectorization, and hardware binding. Different combinations of these decisions produce different candidate schedules and, in turn, different tensor programs. Because the interaction among scheduling decisions is highly combinatorial, tensor-program optimization is commonly formulated as a search problem over candidate schedules.

A practical difficulty is that direct performance measurement on target hardware is expensive. As a result, the compiler usually cannot measure every generated candidate. Instead, it first generates candidate schedules, then evaluates and ranks them, and finally selects only a subset for hardware measurement. Existing work has improved this process from different directions, including search-space design, exploration strategy, and learned candidate evaluation. Our work focuses on the candidate-evaluation stage inside this tensor-program optimization pipeline.

\subsubsection{World Models and Latent State Dynamics}

A world model is a predictive surrogate designed to learn the transition dynamics of an environment. It maps complex, high-dimensional, and often noisy observations into a structured continuous latent space \cite{lecun2022path,bi2026motus,taniguchi2026generative}, where superficial variations are suppressed and semantically meaningful states are represented as distinct geometric coordinates. Instead of reasoning directly over raw observations, a world model learns how latent states evolve under actions, allowing future states to be simulated through compact representation-space rollout.

The central abstraction of a world model is therefore not only state prediction, but also \emph{structured latent state evolution}. Given a current observation or state $s_t$, an encoder maps it into a latent representation $z_t = E(s_t)$. Given an action $a_t$, a transition model predicts the next latent state,
\begin{equation}
    z_{t+1} = F(z_t, a_t),
\end{equation}
so that multi-step consequences can be estimated by repeatedly applying the learned transition function. This formulation is particularly useful when direct interaction with the real environment is expensive, or when raw observations contain irrelevant surface-level variations that obscure the underlying semantic state.

This perspective naturally matches tensor-program optimization. The compiler together with the target hardware can be viewed as a dynamic environment, a TensorIR program corresponds to the environment state, and a scheduling transformation corresponds to an action that mutates the current program state. A candidate schedule is therefore not merely a final static code artifact, but the terminal state induced by rolling out a sequence of scheduling actions from an initial program. From this perspective, candidate evaluation can be formulated as learning the latent dynamics of compiler state transitions and using the predicted terminal-state representation to estimate candidate quality before expensive hardware measurement.

\subsection{Motivation}

Candidate schedules in tensor-program search are produced by sequentially applying scheduling actions to an initial program. Because each action mutates the loop nest, scheduling is inherently a step-by-step transformation process where the physical effect of any decision depends on the context established by prior decisions. 

To evaluate such sequential processes, the most straightforward approach is to ignore the history and assess the candidate using only its final, static program state $s_T$, such as the final TensorIR text. However, evaluating a program solely as a static snapshot ignores the transformation semantics and makes the model highly sensitive to syntax variations. For example, consider optimizing a matrix multiplication loop of size $N$. If $N=1024$, applying a loop-tiling action with a factor of $16$ produces a clean nested loop structure. Conversely, if $N=1023$, the same tiling action forces the compiler to generate additional boundary-handling loops (referred to as loop peeling) to handle the remaining elements. Although the core computation and cache reuse behaviors of these two programs are virtually identical, their final TensorIR texts differ significantly due to the extra conditional branches. A static code encoder operating solely on the final code snapshot $s_T$ can easily misclassify these semantically equivalent candidates because it is easily misled by such syntactic noise.

\begin{figure}[t]
    \centering
    \includegraphics[width=\linewidth]{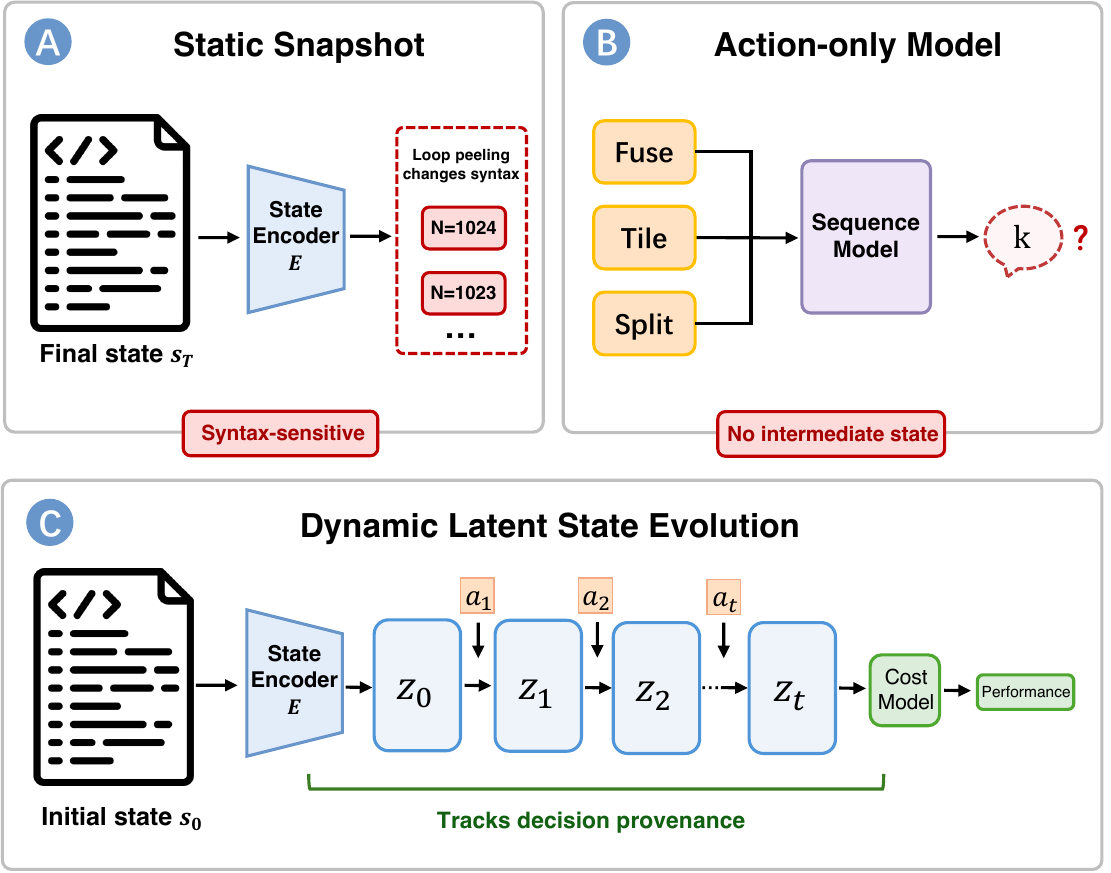}
    \caption{Dynamic latent-state schedule evaluation. The proposed model represents scheduling as action-conditioned latent state evolution, reducing syntax sensitivity and resolving the referential ambiguity of action-only models.}
    \label{fig:static_vs_dynamic}
\end{figure}

To bypass this syntax-level sensitivity, intuitive alternatives include shifting the modeling focus to the scheduling-action sequence $a_{1:T}$ directly (using sequence models like LSTMs or Transformers), or concatenating the final static program state $s_T$ with the action sequence $a_{1:T}$ (as illustrated in Figure~\ref{fig:static_vs_dynamic}). However, both approaches remain fundamentally static and context-blind regarding the intermediate steps. In compilers, scheduling actions are not independent, self-contained tokens; instead, they are referential operators whose targets are dynamically created, renamed, or destroyed. For example, an initial action might fuse two adjacent loop variables $i$ and $j$ to create a new loop variable $k$, which is then tiled in a subsequent step. If an evaluator simply processes the action sequence $[\text{Fuse}(i, j), \text{Tile}(k, 16)]$, either in isolation or concatenated with the final static code $s_T$, it cannot resolve what the entity $k$ actually represents, how large it is, or what its memory properties are. This is because the physical properties and existence of $k$ are strictly defined within the intermediate program state $s_1$ that existed immediately after fusion. Because the intermediate program states $s_1, \dots, s_{T-1}$ are entirely omitted, the evaluator is forced to solve a non-linear inverse problem, attempting to reconstruct the sequential state-action interactions retrospectively from flat action tokens and the end-state code $s_T$. The state-action interaction gap and the referential ambiguity thus remain unresolved, as the step-wise intermediate contexts are never explicitly represented.

This fundamental limitation of static concatenations indicates that schedule evaluation requires a dynamic, action-conditioned state evolution, as illustrated in Figure~\ref{fig:static_vs_dynamic}. We propose to model this process by mapping the initial program state to a latent representation $z_0 = E(s_0)$, and then simulating each scheduling action as an algebraic transition operator that updates the representation step-by-step ($z_{t-1} \xrightarrow{a_t} z_t$). Under this formulation, when the action $\text{Tile}(k)$ is applied, the transition model directly mutates the current latent loop representation $z_{t-1}$ (which already encodes the fused loop $k$) to produce the fused-and-tiled state representation $z_t$. By explicitly coupling the actions with the intermediate states they mutate, this dynamic approach tracks the decision provenance of the transformations, resolving the referential gap of action-only models, the state-action gap of static concatenations, and the syntactic sensitivity of static models.

Furthermore, by simulating these transition dynamics in a continuous latent space, our world-model-inspired framework aims to capture the step-wise physical interactions of scheduling decisions while avoiding the heavy computational overhead of performing intermediate AST mutations inside the compiler during active search.

\begin{figure*}[t]
    \centering
    \includegraphics[width=0.95\linewidth]{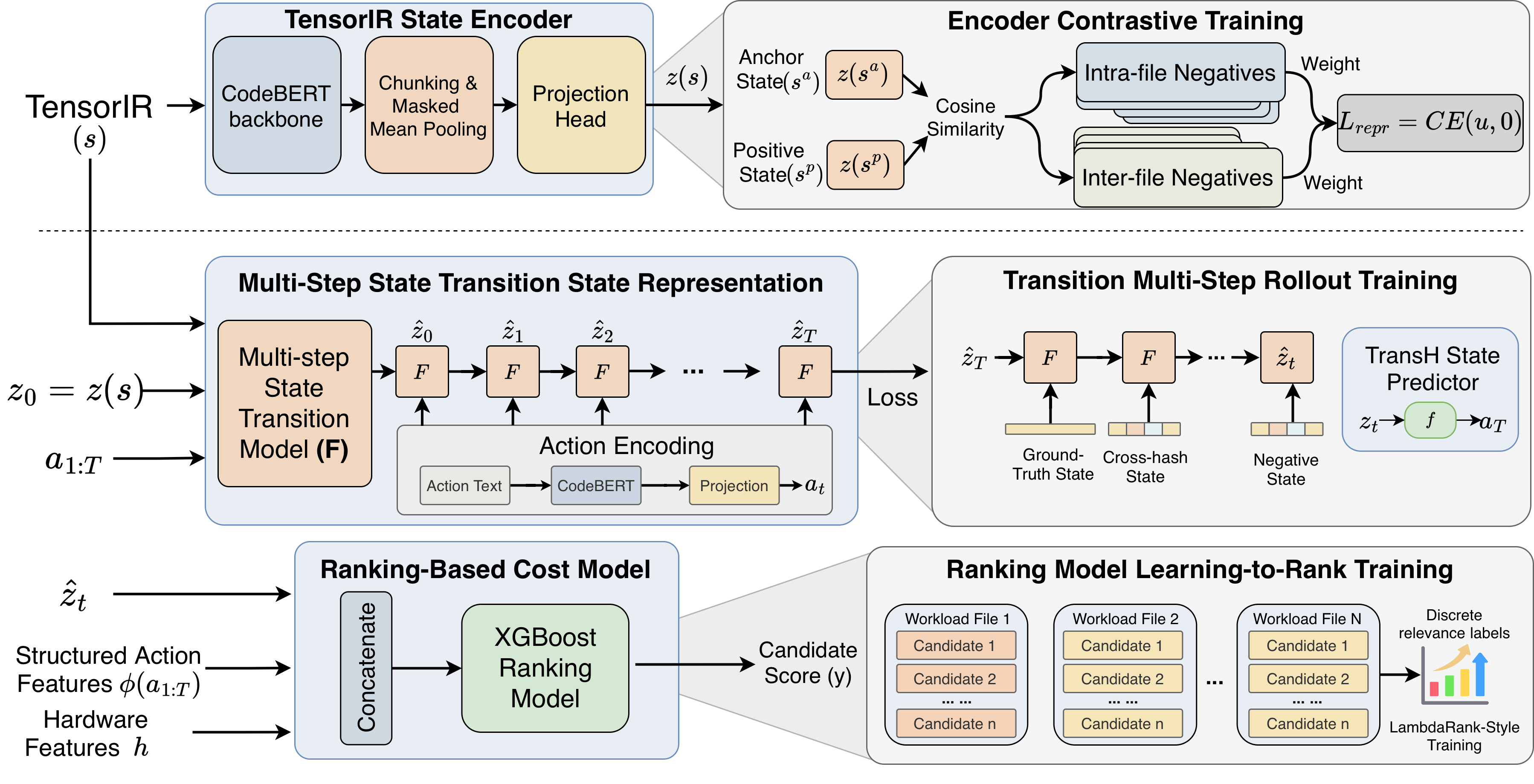}
    \caption{Overview of our framework. Given an initial TensorIR state and a candidate scheduling-action sequence, the framework evaluates the candidate through the terminal-state representation induced by that sequence. Starting from the initial program state, it models action-conditioned state evolution in representation space and uses the predicted terminal-state representation together with action and hardware features to score candidate schedules. The resulting scores are used to guide candidate selection in the online search loop of TVM AutoScheduler.}
    \label{fig:overview}
\end{figure*}

\section{Method}
\label{sec:method}
\subsection{Overview}
\label{sec:overview}

We study schedule evaluation in tensor-program search. Given an initial tensor program, a candidate scheduling-action sequence, and optional hardware information, the goal is to estimate the quality of the resulting candidate before expensive on-device measurement. Our framework models this process in three stages. It first maps a TensorIR program state to a dense representation, then applies an action-conditioned multi-step state transition model to predict the terminal-state representation induced by the scheduling-action sequence, and finally uses the predicted terminal-state representation together with action-structure and hardware features in a ranking-based cost model. The output is a score used to rank candidate schedules within the same workload rather than a direct replacement for measured runtime. In the current implementation, these stages correspond to a TensorIR encoder, a multi-step state predictor, and a ranking cost model.

This decomposition separates three parts of schedule evaluation that are otherwise entangled in a single end-to-end predictor: representing tensor-program states, modeling how scheduling actions change the state, and ranking the resulting candidates. Figure~\ref{fig:overview} illustrates the overall framework. We instantiate it in TVM AutoScheduler, where the ranking scores are used to guide candidate selection in the online search loop. The following subsections describe the formulation and implementation of these components.

\subsection{Problem Formulation}
\label{sec:problem}

We consider tensor-program search for a fixed workload. Let $s_0$ denote the initial tensor-program state before scheduling, and let $a_{1:T} = (a_1, a_2, \dots, a_T)$ denote a candidate scheduling-action sequence of length $T$. Applying $a_{1:T}$ to $s_0$ yields a candidate program state $s_T$, which can be compiled and measured on target hardware. Since such measurement is expensive, the search procedure typically evaluates many candidate action sequences with a learned score and measures only a small subset. Our goal is therefore to learn a scoring function that estimates the relative quality of the candidate induced by $(s_0, a_{1:T})$ before hardware measurement.

We decompose this scoring process into three functions. First, a state encoder $E(\cdot)$ maps a tensor-program state to a dense representation. Second, a transition model $F(\cdot)$ predicts the terminal-state representation induced by the action sequence through multi-step state transition, i.e.,
\begin{equation}
\hat{\mathbf{z}}_T = F(E(s_0), a_{1:T}),
\end{equation}
where $\hat{\mathbf{z}}_T$ denotes the predicted representation of the terminal state. Third, a ranking model $G(\cdot)$ maps the predicted terminal-state representation, together with action-structure features $\phi(a_{1:T})$ and optional hardware features $\mathbf{h}$, to a scalar score,
\begin{equation}
y = G(\hat{\mathbf{z}}_T, \phi(a_{1:T}), \mathbf{h}).
\end{equation}
The score $y$ is used to rank candidate schedules within the same workload. In this work, $y$ is not interpreted as a direct runtime prediction; instead, it is used for relative candidate prioritization during search.

\subsection{Tensor Program State Representation}
\label{sec:state-repr}

The first component of the framework is a representation function for tensor-program states. Since the later stages in the pipeline operate on program states rather than directly on measured runtimes, we first map each TensorIR state to a dense vector representation that can be reused by the state-transition and ranking modules. In the current system, a state is represented by its TensorIR text extracted from collected tuning records, including both pre-schedule and post-schedule program states. The representation model is trained separately before the transition and ranking stages and is then reused as a shared state encoder in the rest of the pipeline.

We implement the encoder using a trainable CodeBERT-based \cite{feng2020codebert} architecture adapted to long TensorIR inputs. Given a TensorIR program $s$, we first tokenize its text into a sequence of tokens. Since TensorIR inputs can exceed the standard Transformer input length, we partition the token sequence into multiple overlapping chunks using a sliding window. Each chunk is encoded independently by the backbone model, and masked mean pooling is applied to obtain a chunk-level representation. These chunk representations are then aggregated into a single program-level vector and passed through a projection head to obtain the final state embedding
\begin{equation}
\mathbf{z}(s) = E(s) \in \mathbb{R}^{d}.
\end{equation}

To train the encoder, we use contrastive learning \cite{simCLR}. At a high level, contrastive learning trains the model to place similar samples close to each other in the representation space while pushing dissimilar samples apart. In our setting, the goal is not to predict runtime directly at this stage, but to construct a state space in which different TensorIR states can be compared by their learned embeddings. The training data are organized at the file level, where each file contains multiple post-schedule TensorIR states associated with the same workload source. For each anchor state, the positive example is constructed from the same TensorIR text, yielding two stochastic views of the same state representation during training. Negative samples are divided into two groups. The first group contains \emph{intra-file negatives}, sampled from other post-schedule TensorIR states in the same file as the anchor. These negatives correspond to different schedule instances associated with the same workload source. The second group contains \emph{inter-file negatives}, sampled from TensorIR states taken from different files.  This yields a training tuple of the form
\begin{equation}
\bigl(s^{a},\, s^{p},\, \mathcal{N}_{\mathrm{intra}}(s^{a}),\, \mathcal{N}_{\mathrm{inter}}(s^{a})\bigr),
\end{equation}
where $s^{a}$ and $s^{p}$ denote the anchor and positive states, and $\mathcal{N}_{\mathrm{intra}}$, $\mathcal{N}_{\mathrm{inter}}$ denote the two negative sets. The two negative groups are kept separate in the loss so that they can be assigned different weights during training.

Let $\mathbf{u}$ denote the logit vector formed by one positive term and the weighted intra-file and inter-file negative terms. We minimize
\begin{equation}
\mathcal{L}_{\mathrm{repr}} = \mathrm{CE}(\mathbf{u}, 0),
\end{equation}
where the positive logit is given by the cosine similarity between the anchor and positive embeddings, and the negative logits are given by the weighted cosine similarities between the anchor and the two groups of negatives. After contrastive training, the encoder provides a mapping from TensorIR states to dense vectors, and these vectors are used as the input state representations for the action-conditioned multi-step transition model described next.

\subsection{Action-Conditioned Multi-Step State Transition}
\label{sec:state-transition}

Given the state representation described above, the next component models how a tensor-program state evolves under a scheduling-action sequence. The purpose of this stage is not to reconstruct TensorIR text, but to provide a terminal-state representation for downstream candidate evaluation. Starting from the embedding of an initial state, the model rolls out the effect of scheduling actions step by step in representation space and outputs the predicted embedding of the terminal state used by the downstream ranking model.

Training data are constructed from tuning logs together with aligned TensorIR states. From each raw record, we extract a pre-schedule state, a post-schedule state, and the corresponding scheduling-action sequence. We then group records by trajectory hash and reconstruct them into action-state trajectories. For a trajectory
\begin{equation}
(s_0, a_1, s_1, a_2, s_2, \dots, a_K, s_K),
\end{equation}
we construct fixed-length rollout windows. Each window contains the current state, the subsequent ground-truth future states, the aligned action sequence, and a set of negative future states for each rollout step. In the current implementation, step-wise negatives are drawn from two sources: cross-hash negatives sampled from trajectories with different hashes, and same-hash negatives sampled from other post states associated with the same hash. Each training sample therefore contains a current state, a multi-step action sequence, aligned future states, rollout masks, and per-step negative states.

To represent actions, we serialize each full action tuple into text and encode it with a frozen CodeBERT action encoder, followed by a trainable projection layer. This produces a dense action embedding for transition modeling. Let $\mathbf{z}_{t-1}$ denote the current state embedding, and let $\mathbf{a}_t$ denote the projected dense feature of the $t$-th action. The next-state embedding is predicted through an action-conditioned transformation
\begin{equation}
\hat{\mathbf{z}}_{t} = F(\mathbf{z}_{t-1}, \mathbf{a}_t),
\end{equation}
where $F(\cdot)$ is implemented by a TransH-based \cite{TransH} predictor. TransH originates from knowledge graph representation learning \cite{transE,TransD,TransR}, where it models a relation as a translation performed in a relation-specific hyperplane. We use this formulation as a lightweight geometric transition model: each scheduling action defines a transformation direction that maps the current state embedding to the next one. Concretely, the current state is first transformed in an action-conditioned geometric space, then combined with the action embedding through a residual multilayer transformation, and finally normalized to obtain the predicted next-state embedding. This transition is performed entirely in representation space. The dense action embedding used here is specific to transition modeling and is distinct from the structured action features used later by the ranking model.

Training uses multi-step rollout rather than isolated one-step prediction. Starting from the embedding of the current state, the predictor is applied recurrently over the action sequence, so that the predicted state at one step becomes the input state for the next. At each rollout step, the predicted state is compared with the corresponding ground-truth future state and a set of negative states.
The training objective contains two terms. The first is a state-matching term that pulls the predicted state toward the ground-truth future state in the representation space:
\begin{equation}
\mathcal{L}_{\mathrm{match}}^{(t)} = \|\hat{\mathbf{z}}_t - \mathbf{z}_t\|_2^2.
\end{equation}
The second is a margin-based ranking term that encourages the predicted state to stay closer to the ground-truth future state than to negative states:
\begin{equation}
\mathcal{L}_{\mathrm{rank}}^{(t)} =
\max\!\bigl(0,\; d(\hat{\mathbf{z}}_t,\mathbf{z}_t) - d(\hat{\mathbf{z}}_t,\mathbf{z}_t^{-}) + \gamma \bigr),
\end{equation}
where $\mathbf{z}_t^{-}$ denotes the hardest negative state at step $t$, and $\gamma$ is a margin parameter.
The final rollout loss accumulates these two terms over valid rollout steps:
\begin{equation}
\mathcal{L}_{\mathrm{trans}}
=
\sum_{t=1}^{K} m_t
\left(
\lambda \,\mathcal{L}_{\mathrm{match}}^{(t)}
+
(1-\lambda)\,\mathcal{L}_{\mathrm{rank}}^{(t)}
\right),
\end{equation}
where $m_t$ is the rollout mask at step $t$, and $\lambda$ controls the relative weight between the two terms. In the current implementation, both terms are computed in the normalized state-representation space.

After training, the transition model maps an initial state embedding and a scheduling-action sequence to a predicted terminal-state embedding, which is then used by the ranking model described in the next subsection.

\subsection{Ranking-Based Candidate Evaluation}
\label{sec:ranking}

The final learned component in our framework is a ranking model for candidate evaluation. Its role is to assign a score to each candidate induced by an initial tensor program and a scheduling-action sequence, so that candidates within the same workload can be prioritized before expensive hardware measurement. In our formulation, the model does not directly predict runtime. Instead, it produces a relative score for within-workload candidate ranking, which matches the role of the evaluator in tensor-program search.

For a candidate action sequence $a_{1:T}$ applied to an initial state $s_0$, the ranking model takes three types of input features. The first is the predicted terminal-state representation $\hat{\mathbf{z}}_T$ produced by the multi-step transition model. This feature summarizes the program state predicted after rolling out the action sequence from the initial state. The second is an action feature vector $\phi(a_{1:T})$ extracted from the action sequence itself. In the current implementation, this feature is distinct from the dense action embedding used in the transition model; instead, it is a structured action feature vector derived from the scheduling actions, including statistics such as action counts, sequence length, and action-specific numeric attributes. The third is an optional hardware feature vector $\mathbf{h}$, obtained from the target hardware description when hardware information is enabled. These features are concatenated and passed to the ranking model:
\begin{equation}
y = G\!\left(\hat{\mathbf{z}}_T, \phi(a_{1:T}), \mathbf{h}\right),
\end{equation}
where $y$ is the scalar score used for candidate ranking. In the current system, $G(\cdot)$ is implemented as an XGBoost \cite{chen2016xgboost} ranking model.

Training data for the ranking model are constructed from measured tuning records. Each sample contains an initial TensorIR state, an action sequence, and the measured execution time of the resulting candidate. Rather than fitting measured runtime directly as a regression target, we organize samples by workload and treat each workload as a ranking group. Within each group, candidates are sorted by measured runtime from fast to slow, and the sorted order is mapped to discrete relevance labels. Measured runtime is therefore used to define within-workload ranking supervision rather than direct runtime regression targets. 

We train the ranking model with a learning-to-rank objective. In the current implementation, this objective is instantiated with LambdaRank \cite{LambdaRank} training through XGBoost using workload-level groups. At inference time, the model outputs a score for each candidate, and the search procedure uses these scores to prioritize candidates for further hardware measurement.

\subsection{Online Integration}
\label{sec:online}

After training the three learned components described above, we integrate them into the online search loop of a tensor-program tuning system. In the current implementation, we instantiate this integration in TVM AutoScheduler by replacing the default candidate-scoring pathway with the learned evaluator. 

For each candidate state proposed during search, the online evaluator first extracts the associated scheduling-action sequence. It then takes the initial tensor-program state of the corresponding task as the starting point, encodes this state using the pretrained TensorIR encoder, and rolls out the action sequence through the multi-step transition model to obtain a predicted terminal-state representation. In parallel, the evaluator extracts the structured action feature vector from the same action sequence and, when enabled, appends the hardware feature vector associated with the target platform. These features are combined and passed to the ranking model, which outputs a scalar score for the candidate. The search procedure uses this score to rank candidate states and prioritize which candidates should be measured on target hardware. In this sense, the learned evaluator serves as a drop-in scoring component inside the existing search loop rather than as a separate search algorithm. 

The role of the online integration is therefore to connect the learned state representation, action-conditioned state transition, and ranking-based candidate evaluation modules with the candidate generation and measurement stages already provided by the tensor-program search framework. This allows us to evaluate the learned pipeline not only through offline metrics, but also through its effect on practical search behavior in a working compiler system.

\section{Experiments}
\subsection{Experimental Setup}
\label{sec:setup}

\begin{table}[t]
\small
\centering
\caption{Benchmark workloads used in our evaluation.}
\label{tab:benchmarks}
\resizebox{\linewidth}{!}{
\begin{tabular}{l l c}
\toprule
Model / Subgraph Source & Input Shape & \# Extracted Subgraphs \\
\midrule
ResNet-50 & $[1, 3, 224, 224]$ & 27 \\
BERT-base & $[1, 128, 768]$ & 8 \\
MobileNetV2 & $[1, 3, 224, 224]$ & 32 \\
GPT-2 & $[1, 128, 768]$ & 9 \\
ResNet18-3D & $[1, 3, 16, 112, 112]$ & 16 \\
OPT-1.3B Self-Attention & $[1, 128, 2048]$ & 6 \\
GPT-Neo-2.7B Self-Attention & $[1, 128, 768]$ & 5 \\
\midrule
Total & -- & 103 \\
\bottomrule
\end{tabular}
}
\end{table}

\begin{table}[t]
\centering
\caption{Representative subgraph types covered in our benchmark set and their corresponding representative tasks used in the evaluation.}
\label{tab:subgraph-types}
\footnotesize
% \resizebox{\linewidth}{!}{
\begin{tabular}{l l}
\toprule
Subgraph Type & Representative Subgraph \\
\midrule
Conv3d+BN+ReLU & ResNet3D-18 task1 \\
Conv3d & ResNet3D-18 task15 \\
FFN & BERT-base task2 \\
Attention (QK MatMul) & OPT-1.3B task1 \\
LayerNorm (variance) & GPT-2 task0 \\
LayerNorm (mean) & GPT-2 task8 \\
Attention (projection) & OPT-1.3B task4 \\
Softmax & OPT-1.3B task3 \\
Fully Connected & ResNet-50 task0 \\
AvgPool2d & MobileNetV2 task1 \\
PV Attention Value Aggregation & BERT-base task6 \\
QKV Projection MatMul & GPT-Neo-2.7B task0 \\
MaxPool2d & ResNet-50 task21 \\
Conv2d + ReLU & ResNet-50 task6 \\
Conv2d & MobileNetV2 task31 \\
DepthConv + ReLU & MobileNetV2 task4 \\
Conv2d + Add & MobileNetV2 task6 \\
DepthConv & MobileNetV2 task22 \\
Conv2d + Add + ReLU & ResNet-50 task7 \\
Conv2d + Pad + ReLU & ResNet-50 task3 \\
Conv2d + Stride + ReLU & ResNet-50 task5 \\
Conv2d + Stride + Add & ResNet-50 task26 \\
\bottomrule
\end{tabular}
% }
\end{table}

\paragraph{Benchmarks.}
We evaluate the proposed method on seven representative workloads covering CNN,
Transformer, and 3D vision settings. Table~\ref{tab:benchmarks} summarizes the
corresponding input shapes and the numbers of extracted subgraphs. In total,
our benchmark set contains 103 extracted subgraphs. These subgraphs cover a
diverse set of operator patterns, including convolution variants, fused residual
blocks, depthwise convolutions, feed-forward networks, attention-related matrix
multiplications, normalization, softmax, fully connected layers, and pooling
operators.

For fine-grained representative-subgraph analysis, we further group the 103
extracted subgraphs by subgraph type and retain one representative task for
each type, yielding 22 representative subgraphs in total. Table~\ref{tab:subgraph-types}
lists these representative subgraph types and their corresponding task
identifiers, so that the task names referenced in later experiments can be
mapped back to the operator categories in the benchmark setup. This
representative subset is used only for per-subgraph analysis, while the
model-level results aggregate over all extracted subgraphs.

\paragraph{Hardware and software.}
All experiments are conducted on a machine equipped with an Intel Xeon Gold 6430 CPU and an NVIDIA GeForce RTX 4090 GPU. Our implementation is built on TVM \texttt{0.8.dev0} with CUDA \texttt{12.6} for GPU execution.

\paragraph{Training data.}
All learned components are trained from tensor programs extracted from
TenSet tuning logs. The contrastive state-representation dataset
contains 10,000 samples, split into 8,000 training samples and 2,000
validation samples. The multi-step state-transition model is trained
separately for CPU and GPU, with 391,240 / 395,761 state-action samples,
yielding 312,992 / 316,608 training samples and 78,248 / 79,153
validation samples after preprocessing. For the ranking-based cost
model, we re-measure the corresponding tensor programs on our own
platforms to obtain up-to-date performance labels, resulting in 140,262
training samples on GPU and 500,425 training samples on CPU.
% \paragraph{Baselines.}
% For the main comparison, we evaluate against three search-based tensor-program tuning baselines: Ansor, FlexTensor, and MetaSchedule. In addition, we compare against TensorRT as a deployment-oriented reference rather than a directly comparable search-based baseline. Unless otherwise noted, all search-based methods are evaluated on the same hardware platform under the same tuning budget.

\subsection{End-to-End Model Performance}
\label{sec:eval-e2e}

We first evaluate the end-to-end performance of the optimized
models.  This experiment compares our method with PyTorch,
PyTorch-opt, TensorRT, and Ansor on the seven benchmark models
described in Section~\ref{sec:setup}.  PyTorch-opt
denotes PyTorch execution with cuDNN-enabled backend optimizations.
TensorRT is included as a vendor-optimized deployment reference,
while Ansor is the most direct comparison because both Ansor and our
method optimize tensor programs within the TVM AutoScheduler pipeline.
All reported values are end-to-end model latencies in milliseconds.

\begin{figure}[t]
  \centering
  \includegraphics[width=0.48\textwidth]{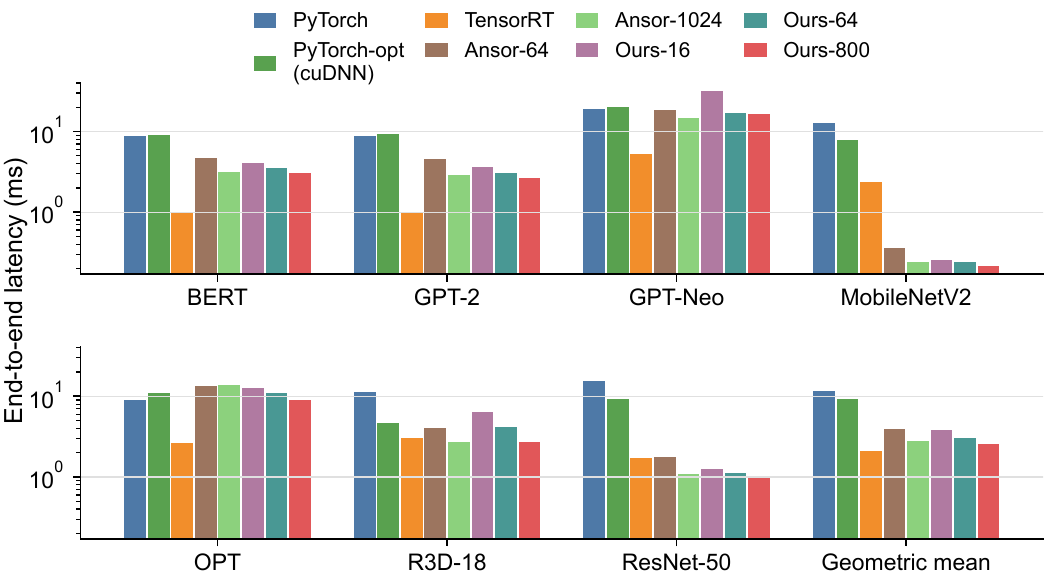}
  \caption{End-to-end latency across seven models. 
  PyTorch-opt denotes cuDNN-enabled PyTorch execution.  The last group
  reports the geometric mean across the seven models.}
  \label{fig:e2e-performance}
  \vspace{-4mm}
\end{figure}

Figure~\ref{fig:e2e-performance} shows that our method improves the
end-to-end performance of the TVM-based AutoScheduler pipeline.  Under
the same 64-trial setting, our method reduces the geometric-mean
latency from 3.91~ms with Ansor-64 to 3.02~ms, corresponding to a
1.29$\times$ speedup across the seven models.  With a larger tuning
budget, our method with 800 trials further reduces the geometric-mean
latency to 2.54~ms, compared with 2.79~ms for Ansor-1024.  This
corresponds to a 1.10$\times$ speedup, and our method
matches or outperforms Ansor-1024 on six of the seven models.

The benefit is visible across different workload families.  Compared
with Ansor-1024, our 800-trial result improves BERT, GPT-2, MobileNetV2,
OPT, ResNet18-3D, and ResNet-50, with the largest gain on OPT
(1.55$\times$).  GPT-Neo is the only model where Ansor-1024 remains
faster in this end-to-end comparison.  These results indicate that the
learned action-conditioned evaluator can improve the schedules selected
by the online search loop and that these improvements translate to
model-level execution latency rather than only isolated kernel metrics.

Compared with framework execution, our method also provides substantial
end-to-end acceleration.  Ours-800 achieves a 4.61$\times$
geometric-mean speedup over PyTorch and a 3.67$\times$ speedup over
PyTorch-opt.  TensorRT remains a strong deployment baseline, especially
for standard Transformer workloads, but it serves a different role: it
is a vendor-optimized inference engine rather than a TVM AutoScheduler
variant.  We therefore use the following experiments to isolate the
search quality of our method against Ansor under matched TVM tuning
settings.

\subsection{Model-Level Search Performance}
\label{sec:eval-model-level}

The end-to-end results include effects from framework runtimes, graph
execution, and deployment engines.  To isolate the quality of tensor
program search itself, we next compare our method with Ansor under the
same TVM AutoScheduler setting.  For each model, we aggregate optimized
subgraph latency using the weighted sum
\begin{equation}
  L_{\mathrm{model}} = \sum_i w_i L_i,
\end{equation}
where \(L_i\) is the best measured latency of subgraph \(i\), and
\(w_i\) is its occurrence count in the model.  This metric is not an
end-to-end runtime measurement; instead, it provides a controlled
model-level view of AutoScheduler search quality.

\begin{figure}[t]
  \centering
  \includegraphics[width=0.48\textwidth]{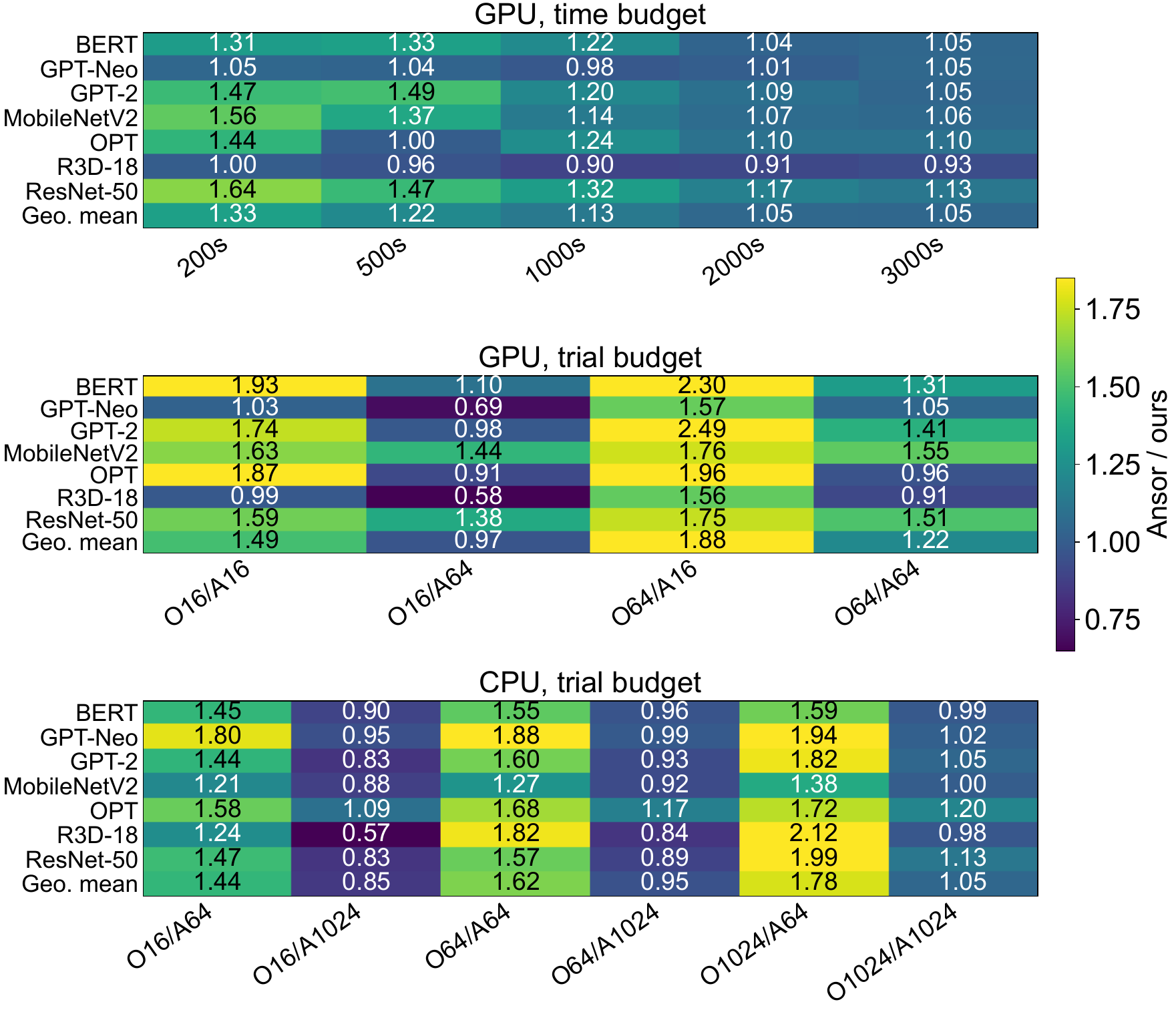}
  \caption{Model-level weighted-latency speedup over Ansor.  Each entry
  reports \(\mathrm{Ansor}/\mathrm{Ours}\), so values larger than one
  indicate that our method obtains lower weighted latency.  The last row
  reports the geometric mean across the seven models.}
  \label{fig:model-level-weighted}
\end{figure}

\begin{figure*}[t]
  \centering
  \includegraphics[width=\textwidth]{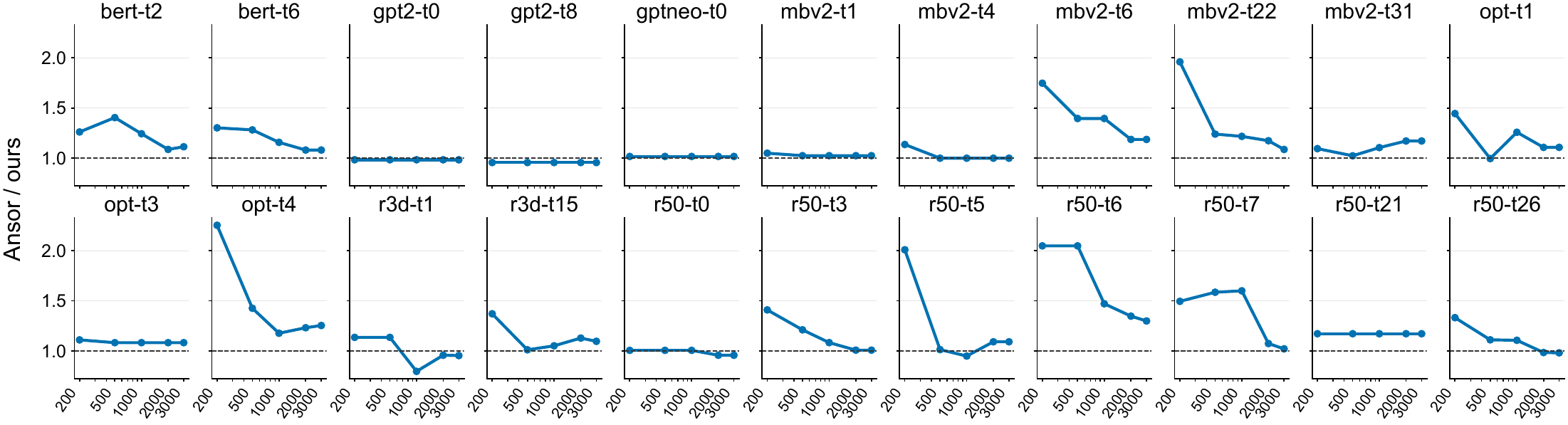}
  \caption{GPU representative-subgraph speedup over time.  Each panel
  shows one representative subgraph, and the y-axis reports
  \(\mathrm{Ansor}/\mathrm{Ours}\).  The time-budget analysis starts
  from 200 seconds.}
  \label{fig:rep-gpu-time-curves}
\end{figure*}

Figure~\ref{fig:model-level-weighted} summarizes the results.  On the
GPU time-budget experiment, our method consistently improves over Ansor
from 200 seconds onward.  The geometric-mean speedup is 1.33$\times$ at
200 seconds, 1.22$\times$ at 500 seconds, and remains above
1.05$\times$ at 3000 seconds.  This shows that the proposed evaluator
is particularly helpful under limited tuning time, while still
maintaining an advantage after longer tuning.

We also compare fixed measurement budgets.  On GPU, our method obtains
a 1.49$\times$ geometric-mean speedup over Ansor at the same 16-trial
budget and a 1.22$\times$ speedup at the same 64-trial budget.  Under a
cross-budget comparison, Ours-64 is 1.88$\times$ faster than Ansor-16,
while Ours-16 is close to Ansor-64.  These results indicate that the
learned evaluator improves the efficiency of the online search loop,
allowing the scheduler to find better candidates with the same or fewer
measurements.

The CPU results show a similar trend.  At the same 64-trial budget, our
method achieves a 1.62$\times$ geometric-mean speedup over Ansor.  At
1024 trials, the speedup is smaller but remains positive at
1.05$\times$.  In the cross-budget setting, Ours-16 already outperforms
Ansor-64 by 1.44$\times$ on average, while Ours-64 nearly matches
Ansor-1024.  Together with the GPU results, this demonstrates that the
benefit of action-conditioned candidate evaluation is not limited to a
single hardware target.

The model-level weighted-latency results support the conclusion
from the end-to-end experiment: our method improves the schedules found
by the TVM AutoScheduler pipeline.  The improvement is strongest under
small search budgets, which is the regime where candidate evaluation has
the largest impact because only a limited number of schedules can be
measured on hardware.

\subsection{Representative Subgraph Analysis}
\label{sec:eval-representative-subgraphs}

To understand where the model-level improvements come from, we further
evaluate the 22 representative subgraphs used in our analysis.  This
experiment removes model-level aggregation and compares Ansor and our
method at the individual subgraph level.  As in
Section~\ref{sec:eval-model-level}, speedup is reported as
\(\mathrm{Ansor}/\mathrm{Ours}\), so values larger than one indicate
that our method obtains a lower measured latency.

Figure~\ref{fig:rep-gpu-time-curves} shows how the relative performance
of the 22 representative subgraphs changes as the tuning time increases.
The curves show that the advantage of our method is strongest under
small time budgets and gradually narrows as both methods are given more
time to search.  Aggregated across the 22 subgraphs, our method achieves
a 1.33$\times$ geometric-mean speedup over Ansor at 200 seconds,
1.17$\times$ at 500 seconds, 1.12$\times$ at 1000 seconds, and
1.07$\times$ at 3000 seconds.  This trend is consistent with the
model-level time-budget results in Section~\ref{sec:eval-model-level}:
action-conditioned evaluation is most useful when the scheduler must
prioritize candidates under a limited measurement budget.

\input{tables/experiment3_gpu_representative_tenset_baseline.tex}

We also compare against TenSet as an additional learning-based baseline
in the GPU representative-subgraph trial setting. Table~\ref{tab:rep-gpu-tenset} reports the
per-subgraph speedup breakdown at 16, 64, and 1024 trials.  On the
geometric mean, our method is 2.14$\times$ faster than Ansor and
2.36$\times$ faster than TenSet at 16 trials.  The speedups remain
1.37$\times$ over Ansor and 1.79$\times$ over TenSet at 64 trials, and
1.09$\times$ over Ansor and 1.32$\times$ over TenSet at 1024 trials.
The per-subgraph entries show that the gap to TenSet is especially
large on several convolutional workloads, while the comparison to Ansor
becomes tighter at 1024 trials but still favors our method on the
geometric mean.

\begin{figure}[t]
  \centering
  \includegraphics[width=0.5\textwidth]{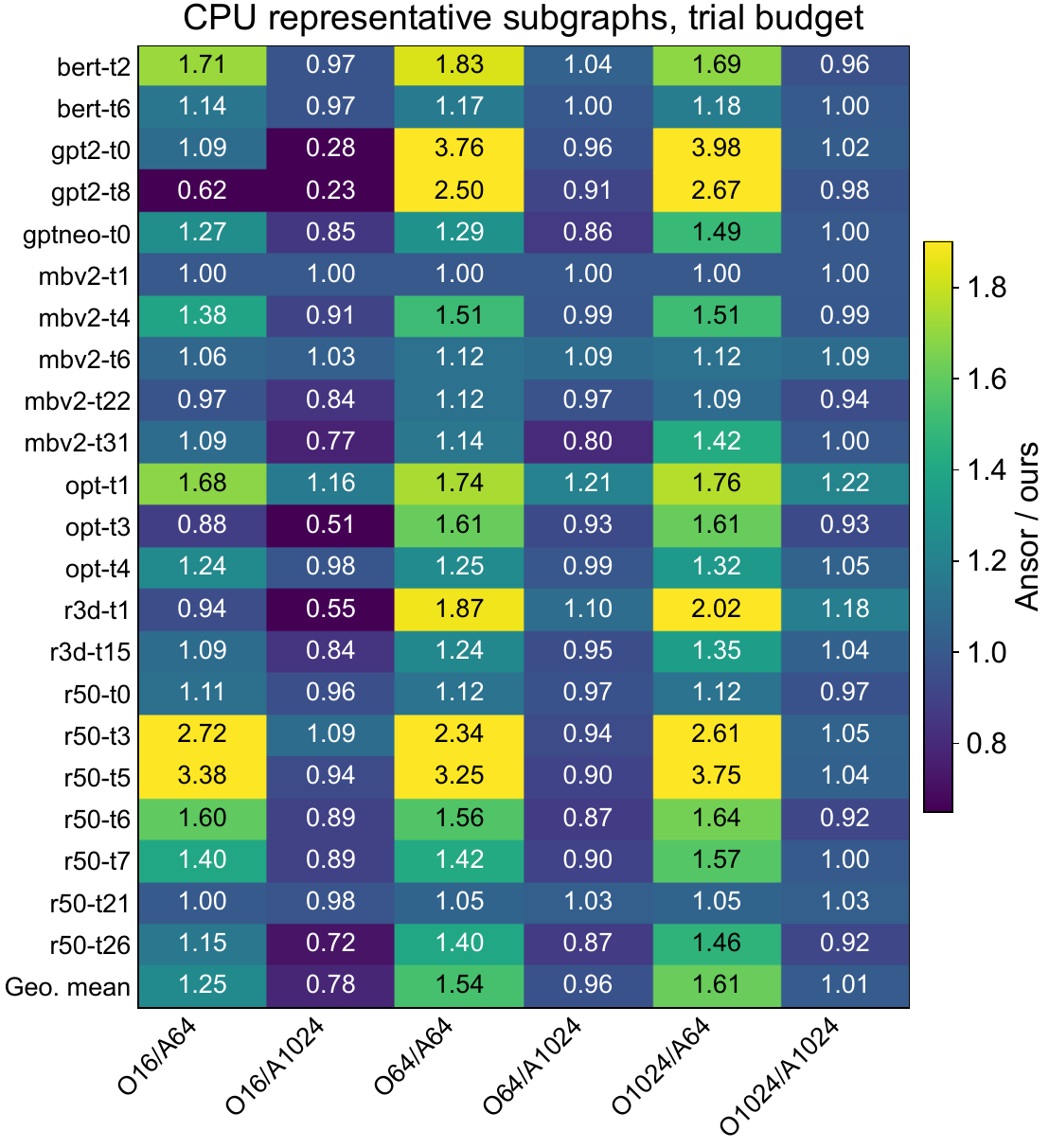}
  \caption{CPU representative-subgraph speedup under cross-trial
  budgets.  Each entry reports \(\mathrm{Ansor}/\mathrm{Ours}\).}
  \label{fig:rep-cpu-trial-heatmap}
  \vspace{-2mm}
\end{figure}

Figure~\ref{fig:rep-cpu-trial-heatmap} shows the corresponding CPU
trial-budget results.  The same-budget comparison again favors our
method: it achieves a 1.54$\times$ geometric-mean speedup at 64 trials
and remains slightly faster at 1024 trials with a 1.01$\times$ speedup.
The cross-budget setting further shows that Ours-16 already outperforms
Ansor-64 by 1.25$\times$, while Ours-64 nearly reaches Ansor-1024.  The
consistent trends on both GPU and CPU indicate that the benefit of our
action-conditioned evaluator is not tied to a single hardware target or
to a small number of subgraphs.

The representative-subgraph analysis explains the source of the
model-level gains in Section~\ref{sec:eval-model-level}.  Our method
improves search quality for many individual subgraphs, especially under
small time or measurement budgets, and these subgraph-level improvements
accumulate into lower weighted model latency.

\subsection{Sample Efficiency vs. Large-Budget Ansor}
\label{sec:eval-sample-efficiency}

The previous experiments show that our method improves search quality
under matched budgets.  We now ask a stronger question: how close can
our method get to substantially larger Ansor tuning budgets?  We compare
our method with 16, 64, and 1024 trials against Ansor with 1024, 2048,
4096, 8192, and 10240 trials on the 22 GPU representative subgraphs.  The
reported ratio is \(\mathrm{Ansor}/\mathrm{Ours}\).  Therefore, values
larger than one indicate that our method is faster than the
corresponding large-budget Ansor run, while values below one indicate
that our method is slower but may still be close to the large-budget
Ansor result.

\begin{figure}[t]
  \centering
  \includegraphics[width=0.5\textwidth]{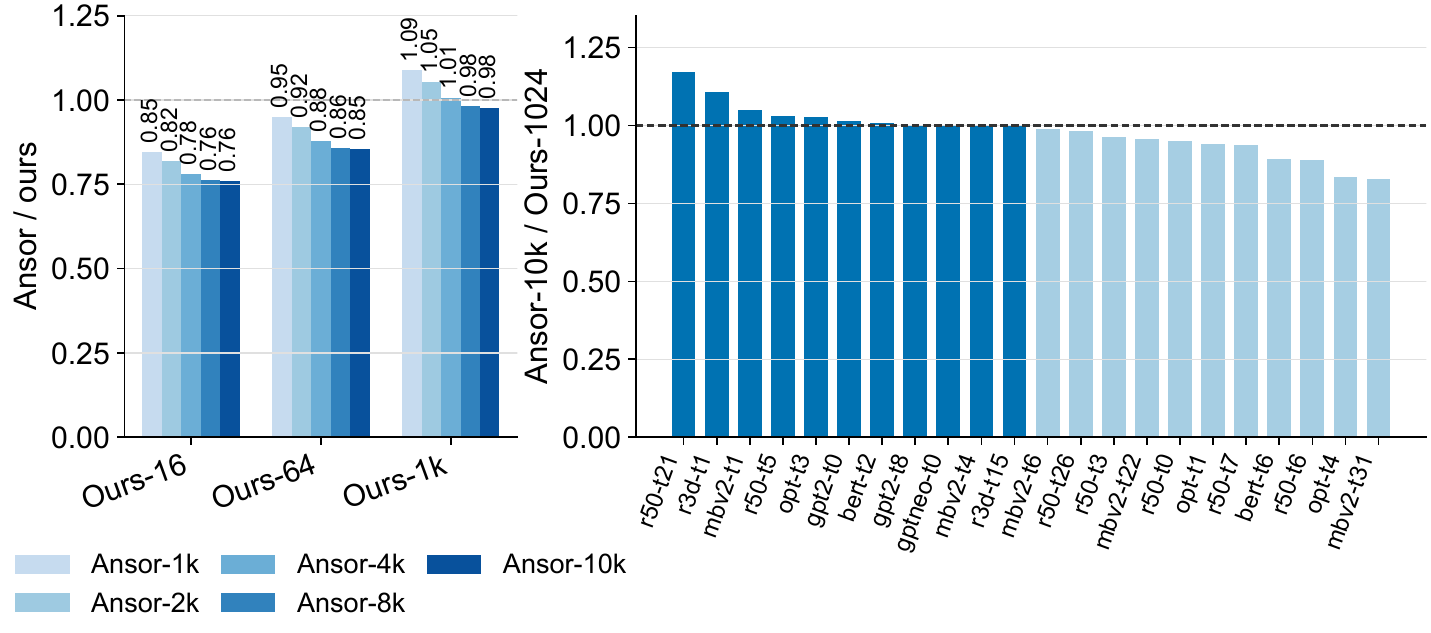}
  \caption{Sample efficiency compared with large-budget Ansor on GPU
  representative subgraphs.  (a) Geometric-mean ratios against
  Ansor-1024, Ansor-2048, Ansor-4096, Ansor-8192, and Ansor-10240.  (b) Per-subgraph
  ratios for Ours-1024 against Ansor-10240, sorted by speedup.}
  \label{fig:sample-efficiency}
  \vspace{-4mm}
\end{figure}

Figure~\ref{fig:sample-efficiency}(a) shows that our method approaches
the quality of much larger Ansor search budgets with substantially fewer
measurements.  With only 64 trials, our method reaches 0.95$\times$ of
Ansor-1024, 0.92$\times$ of Ansor-2048, and 0.88$\times$ of Ansor-4096
on the geometric mean.  With 1024 trials, our method exceeds
Ansor-1024 by 1.09$\times$ and Ansor-2048 by 1.05$\times$, matches
Ansor-4096 with a 1.01$\times$ ratio, and remains close to Ansor-8192
and Ansor-10240 with ratios of 0.98$\times$ and 0.98$\times$,
respectively.  Against Ansor-10240, Ours-1024 matches or outperforms 11
out of 22 representative subgraphs.

Figure~\ref{fig:sample-efficiency}(b) breaks down the Ours-1024 result
by subgraph.  The sorted bars show that the near-parity with Ansor-10240
is not caused by a single extreme outlier: roughly half of the
representative subgraphs are at or above the large-budget Ansor result,
and many of the remaining subgraphs are close to the parity line.  This
supports the conclusion that the learned evaluator improves sample
efficiency by guiding the online search toward high-quality candidates
with fewer hardware measurements.

This experiment complements the matched-budget comparisons in
Sections~\ref{sec:eval-model-level} and
\ref{sec:eval-representative-subgraphs}.  While our method does not
strictly dominate Ansor-10240 at every subgraph, it nearly matches the
geometric-mean performance of that large-budget baseline using only
one-tenth of the trials, and it outperforms it on half of the
representative subgraphs.

\subsection{Ablation Study}
\label{sec:eval-ablation}

We next evaluate the contribution of the main components in our
schedule evaluator.  The ablation is conducted on the 22 representative
subgraphs under a 64-trial search budget.  The full method in this
comparison is also evaluated with the same 64-trial budget, so the
numbers isolate the effect of changing the evaluator rather than the
effect of using more measurements.  For each ablated variant, we report
the latency increase relative to the full method:
\begin{equation}
  \Delta = \frac{L_{\mathrm{ablated}} - L_{\mathrm{full}}}
                 {L_{\mathrm{full}}} \times 100\%.
\end{equation}
Lower values are better, and a positive value means that removing or
changing the component makes the selected schedule slower than the full
method.

We consider five ablated variants.  The first two variants remove the
learned action-conditioned state transition, but differ in which state
representation is used by the cost model.  \emph{w/o transition
(initial state)} feeds the initial TensorIR embedding directly to the
cost model.  \emph{w/o transition (final state)} instead embeds the
compiler-produced final TensorIR state after applying the schedule,
rather than using the learned transition module.  \emph{w/o action
features} removes the structured scheduling-action features from the
cost-model input.  \emph{w/o program state} removes the whole-program
state representation while keeping the structured action features and
hardware embedding.
The regression objective replaces the group-wise \texttt{rank:ndcg} with \texttt{reg:squarederror} using the continuous $\mathrm{best\_t}/t$ target.

\begin{figure}[t]
  \centering
  \includegraphics[width=0.5\textwidth]{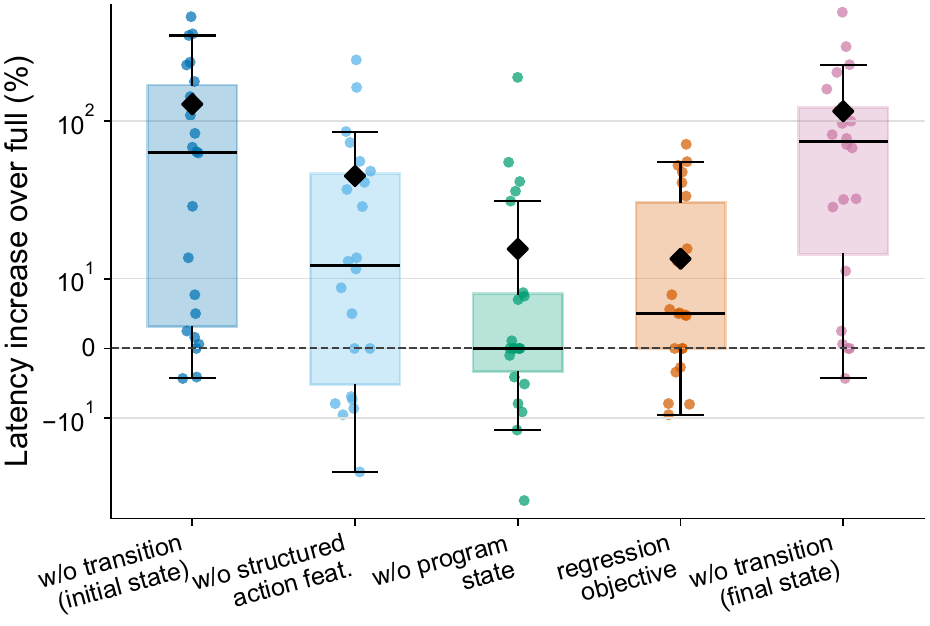}
  \caption{Ablation on 22 subgraphs. Metric is latency increase relative to full method (64 trials). Boxes show distribution across subgraphs, points show individual subgraphs, diamonds indicate means.}
  \label{fig:ablation}
  \vspace{-6mm}
\end{figure}

Figure~\ref{fig:ablation} shows that the learned state-transition module
is the most important component.  Using the initial TensorIR embedding
without transition slows down 19 out of 22 representative subgraphs,
with a median latency increase of 56.5\% and a mean increase of
136.6\%.  Replacing the learned transition with an embedding of the
compiler-produced final state is also substantially worse than the full
method: it slows down 19 subgraphs, with a median increase of 69.0\% and
a mean increase of 120.1\%.  Removing structured action features also
hurts performance, increasing median latency by 12.0\% and slowing down
14 subgraphs.  The \emph{w/o program state} variant has a median
increase of 0.0\%, but its mean increase reaches 14.3\%, indicating
larger regressions on some workloads.  Replacing group-wise ranking with
regression increases median latency by 5.1\% and mean latency by
12.9\%.

These results support the design of the evaluator.  The poor result of
\emph{w/o transition (initial state)} shows that the initial TensorIR
embedding is not sufficient after scheduling; it must be transformed
into a candidate-specific terminal-state representation.  The poor
result of \emph{w/o transition (final state)} further shows that simply
encoding the compiler-produced final state is not enough to replace the
learned action-conditioned transition.  Structured action features
remain useful even with the learned state representation, because they
expose explicit schedule-level decisions such as tiling, binding, and
memory-related transformations to the ranker.  The comparison with
\emph{w/o program state} further suggests that direct action descriptors
can work on some subgraphs, but they do not capture the resulting
program state robustly across workloads.  Finally, the regression
variant shows that the training objective also matters.  Since online
search only needs to rank candidates within the same workload group,
group-wise ranking is better aligned with the selection problem than
fitting a global continuous latency target.

% We also ablate the chunk aggregation strategy used in the long-input
% TensorIR encoder on ResNet-50, using a 16-trial budget over 27
% subgraphs.  The results show a clear front-biased pattern.  Aggregation
% that gives the highest weight to the first chunk performs best, followed
% by using only the first chunk, while averaging the first two chunks
% gives similar but slightly weaker results.  In contrast, uniform mean
% pooling over all chunks performs worst.  These results suggest that
% tuning-relevant information is concentrated primarily in the first
% chunk, while later chunks provide only weak auxiliary context and should
% not be fused with equal weight.

\section{Related Work}

High-performance deep learning execution has long relied on vendor-provided libraries and deployment systems, including cuDNN~\cite{chetlur2014cudnn}, oneDNN~\cite{onednn-docs}, CUTLASS~\cite{cutlass-docs}, and TensorRT~\cite{tensorrt-docs}. These systems provide highly optimized implementations for common operators and deployment scenarios, but they depend on substantial manual engineering and are less flexible when new operators, shapes, or hardware-specific transformations are needed.

Compiler-based systems aim to reduce this dependence by generating optimized tensor programs automatically. TVM established an end-to-end compilation stack for machine learning workloads, with Relay serving as its high-level IR~\cite{chen2018tvm,roesch2018relay}. Similar efforts have also explored extensible compiler infrastructures for machine learning workloads, such as Buddy Compiler, which is built on MLIR and targets both model-level and kernel-level compilation~\cite{zhang2023buddycompiler}.  Tensor Comprehensions and Halide-style systems demonstrated the value of separating computation specification from optimization strategy~\cite{Tensorcomprehensions,Halide}, while FlexTensor and AKG extended this line toward heterogeneous and accelerator-oriented settings~\cite{zheng2020flextensor,zhao2021akg}. In particular, Tensor Comprehensions combines a tensor DSL, a polyhedral JIT compiler, and an autotuner~\cite{Tensorcomprehensions}, and AKG uses polyhedral scheduling to support transformations beyond manually written schedules~\cite{zhao2021akg}. MLIR and Triton further broaden this design space by providing extensible compiler infrastructure and specialized kernel-generation paths for modern accelerators~\cite{lattner2020mlir,lucke2025mlirtransformdialect,tirichine2026reinforcementmlir,tillet2019triton}.

A parallel line of work formulates tensor optimization as search guided by learned evaluators. AutoTVM introduced learning-guided optimization over template-based spaces~\cite{autotvm}. Ansor, MetaSchedule, and related systems expanded the optimization space and improved candidate generation and search control~\cite{zheng2020ansor,MetaSchedule}. TenSet provided a large-scale benchmark and training resource for learned tensor compilers~\cite{zheng2021tenset}. TLP improved learned evaluation by treating schedule primitives as a tensor language and formulating latency prediction as a regression problem~\cite{zhai2023tlp}. Despite these advances, prior cost models score candidates from static program representations, while our work models the intermediate state evolution induced by a scheduling-action sequence.

Recent progress in large language models, exemplified by GPT-4, Claude, Qwen, and DeepSeek~\cite{gpt4,claude3,qwen25,deepseekv3}, has also influenced program optimization. In tensor compilation, TLM uses a tensor language model to guide exploration in large decision spaces~\cite{TLM}. More recent systems explore increasingly autonomous optimization loops, including draft-then-verify tuning, diffusion-based search, instruction-level auto-tuning, and agent-driven kernel optimization~\cite{qiao2025pruner,jeong2025bayesian,ma2025intelligen,jaber2026autokernel,qu2026two}. These methods mainly improve how promising transformation sequences are proposed or refined. Our work is complementary: rather than using a large model or agent to directly generate schedules, we strengthen candidate evaluation by modeling action-conditioned program-state evolution.

\section{Conclusion}
\label{sec:conclusion}

We presented a world-model-inspired framework for tensor program evaluation, adapting latent state-transition modeling to compiler optimization. Rather than evaluating candidate schedules as final static code snapshots, our framework rolls out scheduling transformations as action-conditioned transitions over TensorIR state representations. We also constructed a TVM-based state-prediction dataset that organizes tuning logs and aligned TensorIR states into action-state trajectories for multi-step program-state prediction. Instantiated in TVM AutoScheduler with a TensorIR encoder, a latent state predictor, and a workload-level ranking model, the framework achieves competitive search quality with substantially fewer hardware measurements and delivers significant end-to-end inference speedups over PyTorch baselines. These results suggest that modeling compiler state evolution is a promising direction for efficient tensor program search.

At the same time, this work has several limitations. First, our method
improves candidate evaluation rather than candidate generation or
search-space design. Its effectiveness is still bounded by the coverage
of the underlying scheduler and the quality of the search space it
explores. If strong schedules are not proposed during search, a
stronger evaluator alone cannot recover them. Second, because
terminal-state representations are obtained through multi-step rollout in
latent space, prediction errors may accumulate on long or complex
scheduling trajectories, reducing the reliability of the predicted
terminal state in harder cases. Third, the model is trained for
within-workload candidate ranking rather than absolute latency
prediction. This makes it well suited for online candidate
prioritization, but less directly applicable to settings requiring
calibrated performance estimates or direct comparisons across different
workloads.

\bibliographystyle{ACM-Reference-Format}
\bibliography{sample-base}

\end{document}

%% file: tables/experiment3_gpu_representative_tenset_baseline.tex
\begin{table}[t]
  \centering
  \caption{GPU representative-subgraph speedup against Ansor and TenSet. Each entry reports speedup.}
  \label{tab:rep-gpu-tenset}
  % \scriptsize
  \footnotesize
  \setlength{\tabcolsep}{3.2pt}
  \begin{tabular}{lrrrrrr}
    \toprule
    & \multicolumn{2}{c}{16 trials} & \multicolumn{2}{c}{64 trials} & \multicolumn{2}{c}{1024 trials} \\
    \cmidrule(lr){2-3} \cmidrule(lr){4-5} \cmidrule(lr){6-7}
    Subgraph & A/O & T/O & A/O & T/O & A/O & T/O \\
    \midrule
    bert-t2 & 0.73 & 1.92 & 1.25 & 2.59 & 1.31 & 2.60 \\
    bert-t6 & 1.35 & 2.21 & 2.40 & 2.62 & 0.97 & 1.66 \\
    gpt2-t0 & 13.24 & 12.07 & 1.02 & 1.01 & 1.02 & 0.91 \\
    gpt2-t8 & 3.39 & 3.11 & 1.00 & 0.98 & 1.00 & 0.98 \\
    gptneo-t0 & 15.38 & 14.12 & 1.01 & 1.00 & 1.00 & 0.99 \\
    mbv2-t1 & 1.00 & 1.24 & 1.05 & 1.27 & 1.05 & 1.24 \\
    mbv2-t4 & 1.22 & 2.40 & 1.09 & 2.18 & 1.00 & 1.99 \\
    mbv2-t6 & 3.20 & 2.10 & 1.75 & 2.84 & 1.38 & 1.74 \\
    mbv2-t22 & 1.44 & 4.36 & 1.96 & 4.71 & 1.17 & 2.92 \\
    mbv2-t31 & 2.16 & 1.90 & 1.10 & 1.87 & 1.20 & 1.55 \\
    opt-t1 & 1.29 & 1.69 & 0.92 & 1.76 & 1.01 & 2.17 \\
    opt-t3 & 2.25 & 1.08 & 1.11 & 1.02 & 1.03 & 1.01 \\
    opt-t4 & 1.26 & 2.36 & 2.27 & 2.10 & 1.07 & 1.47 \\
    r3d-t1 & 1.14 & 6.83 & 1.24 & 4.94 & 1.26 & 2.27 \\
    r3d-t15 & 4.02 & 2.37 & 1.66 & 2.44 & 1.15 & 2.05 \\
    r50-t0 & 1.77 & 0.37 & 1.01 & 0.64 & 1.04 & 0.37 \\
    r50-t3 & 1.51 & 3.06 & 1.59 & 2.28 & 1.02 & 1.06 \\
    r50-t5 & 6.63 & 2.03 & 2.01 & 1.02 & 1.09 & 0.86 \\
    r50-t6 & 3.04 & 5.08 & 2.05 & 6.72 & 1.14 & 1.65 \\
    r50-t7 & 0.95 & 5.59 & 1.50 & 5.85 & 0.95 & 4.11 \\
    r50-t21 & 1.10 & 1.07 & 1.17 & 0.99 & 1.17 & 0.99 \\
    r50-t26 & 2.46 & 0.29 & 1.37 & 0.30 & 1.00 & 0.18 \\
    \midrule
    Geometric mean & 2.14 & 2.36 & 1.37 & 1.79 & 1.09 & 1.32 \\
    \bottomrule
  \end{tabular}
  \vspace{-6mm}
\end{table}

%% file: sample-base.bib
@article{chetlur2014cudnn,
  title={cudnn: Efficient primitives for deep learning},
  author={Chetlur, Sharan and Woolley, Cliff and Vandermersch, Philippe and Cohen, Jonathan and Tran, John and Catanzaro, Bryan and Shelhamer, Evan},
  journal={arXiv preprint arXiv:1410.0759},
  year={2014}
}

@inproceedings{chen2018tvm,
  title={$\{$TVM$\}$: An automated $\{$End-to-End$\}$ optimizing compiler for deep learning},
  author={Chen, Tianqi and Moreau, Thierry and Jiang, Ziheng and Zheng, Lianmin and Yan, Eddie and Shen, Haichen and Cowan, Meghan and Wang, Leyuan and Hu, Yuwei and Ceze, Luis and others},
  booktitle={13th USENIX symposium on operating systems design and implementation (OSDI 18)},
  pages={578--594},
  year={2018}
}

@article{Halide,
  title={Learning to optimize halide with tree search and random programs},
  author={Adams, Andrew and Ma, Karima and Anderson, Luke and Baghdadi, Riyadh and Li, Tzu-Mao and Gharbi, Micha{\"e}l and Steiner, Benoit and Johnson, Steven and Fatahalian, Kayvon and Durand, Fr{\'e}do and others},
  journal={ACM Transactions on Graphics (TOG)},
  volume={38},
  number={4},
  pages={1--12},
  year={2019},
  publisher={ACM New York, NY, USA}
}

@article{Tensorcomprehensions,
  title={Tensor comprehensions: Framework-agnostic high-performance machine learning abstractions},
  author={Vasilache, Nicolas and Zinenko, Oleksandr and Theodoridis, Theodoros and Goyal, Priya and DeVito, Zachary and Moses, William S and Verdoolaege, Sven and Adams, Andrew and Cohen, Albert},
  journal={arXiv preprint arXiv:1802.04730},
  year={2018}
}

@inproceedings{zheng2020flextensor,
  title={Flextensor: An automatic schedule exploration and optimization framework for tensor computation on heterogeneous system},
  author={Zheng, Size and Liang, Yun and Wang, Shuo and Chen, Renze and Sheng, Kaiwen},
  booktitle={Proceedings of the Twenty-Fifth International Conference on Architectural Support for Programming Languages and Operating Systems},
  pages={859--873},
  year={2020}
}

@inproceedings{NeonCPU,
  title={Optimizing $\{$CNN$\}$ model inference on $\{$CPUs$\}$},
  author={Liu, Yizhi and Wang, Yao and Yu, Ruofei and Li, Mu and Sharma, Vin and Wang, Yida},
  booktitle={2019 USENIX Annual Technical Conference (USENIX ATC 19)},
  pages={1025--1040},
  year={2019}
}

@inproceedings{abadi2016tensorflow,
  title={$\{$TensorFlow$\}$: a system for $\{$Large-Scale$\}$ machine learning},
  author={Abadi, Mart{\'\i}n and Barham, Paul and Chen, Jianmin and Chen, Zhifeng and Davis, Andy and Dean, Jeffrey and Devin, Matthieu and Ghemawat, Sanjay and Irving, Geoffrey and Isard, Michael and others},
  booktitle={12th USENIX symposium on operating systems design and implementation (OSDI 16)},
  pages={265--283},
  year={2016}
}

@inproceedings{ansel2024pytorch,
  title={Pytorch 2: Faster machine learning through dynamic python bytecode transformation and graph compilation},
  author={Ansel, Jason and Yang, Edward and He, Horace and Gimelshein, Natalia and Jain, Animesh and Voznesensky, Michael and Bao, Bin and Bell, Peter and Berard, David and Burovski, Evgeni and others},
  booktitle={Proceedings of the 29th ACM international conference on architectural support for programming languages and operating systems, volume 2},
  pages={929--947},
  year={2024}
}

@article{chen2015mxnet,
  title={Mxnet: A flexible and efficient machine learning library for heterogeneous distributed systems},
  author={Chen, Tianqi and Li, Mu and Li, Yutian and Lin, Min and Wang, Naiyan and Wang, Minjie and Xiao, Tianjun and Xu, Bing and Zhang, Chiyuan and Zhang, Zheng},
  journal={arXiv preprint arXiv:1512.01274},
  year={2015}
}

@inproceedings{roesch2018relay,
  title={Relay: A new ir for machine learning frameworks},
  author={Roesch, Jared and Lyubomirsky, Steven and Weber, Logan and Pollock, Josh and Kirisame, Marisa and Chen, Tianqi and Tatlock, Zachary},
  booktitle={Proceedings of the 2nd ACM SIGPLAN international workshop on machine learning and programming languages},
  pages={58--68},
  year={2018}
}

@article{autotvm,
  title={Learning to optimize tensor programs},
  author={Chen, Tianqi and Zheng, Lianmin and Yan, Eddie and Jiang, Ziheng and Moreau, Thierry and Ceze, Luis and Guestrin, Carlos and Krishnamurthy, Arvind},
  journal={Advances in Neural Information Processing Systems},
  volume={31},
  year={2018}
}

@inproceedings{zheng2020ansor,
  title={Ansor: Generating $\{$High-Performance$\}$ tensor programs for deep learning},
  author={Zheng, Lianmin and Jia, Chengfan and Sun, Minmin and Wu, Zhao and Yu, Cody Hao and Haj-Ali, Ameer and Wang, Yida and Yang, Jun and Zhuo, Danyang and Sen, Koushik and others},
  booktitle={14th USENIX symposium on operating systems design and implementation (OSDI 20)},
  pages={863--879},
  year={2020}
}

@article{MetaSchedule,
  title={Tensor program optimization with probabilistic programs},
  author={Shao, Junru and Zhou, Xiyou and Feng, Siyuan and Hou, Bohan and Lai, Ruihang and Jin, Hongyi and Lin, Wuwei and Masuda, Masahiro and Yu, Cody Hao and Chen, Tianqi},
  journal={Advances in Neural Information Processing Systems},
  volume={35},
  pages={35783--35796},
  year={2022}
}

@inproceedings{zhao2021akg,
  title={AKG: automatic kernel generation for neural processing units using polyhedral transformations},
  author={Zhao, Jie and Li, Bojie and Nie, Wang and Geng, Zhen and Zhang, Renwei and Gao, Xiong and Cheng, Bin and Wu, Chen and Cheng, Yun and Li, Zheng and others},
  booktitle={Proceedings of the 42nd ACM SIGPLAN International Conference on Programming Language Design and Implementation},
  pages={1233--1248},
  year={2021}
}

@inproceedings{chen2016xgboost,
  title={Xgboost: A scalable tree boosting system},
  author={Chen, Tianqi and Guestrin, Carlos},
  booktitle={Proceedings of the 22nd acm sigkdd international conference on knowledge discovery and data mining},
  pages={785--794},
  year={2016}
}

@inproceedings{zheng2021tenset,
  title={Tenset: A large-scale program performance dataset for learned tensor compilers},
  author={Zheng, Lianmin and Liu, Ruochen and Shao, Junru and Chen, Tianqi and Gonzalez, Joseph E and Stoica, Ion and Ali, Ameer Haj},
  booktitle={Thirty-fifth Conference on Neural Information Processing Systems Datasets and Benchmarks Track (Round 1)},
  year={2021}
}

@inproceedings{zhai2023tlp,
  title={Tlp: A deep learning-based cost model for tensor program tuning},
  author={Zhai, Yi and Zhang, Yu and Liu, Shuo and Chu, Xiaomeng and Peng, Jie and Ji, Jianmin and Zhang, Yanyong},
  booktitle={Proceedings of the 28th ACM International Conference on Architectural Support for Programming Languages and Operating Systems, Volume 2},
  pages={833--845},
  year={2023}
}

@inproceedings{TLM,
  title={Enabling tensor language model to assist in generating $\{$High-Performance$\}$ tensor programs for deep learning},
  author={Zhai, Yi and Yang, Sijia and Pan, Keyu and Zhang, Renwei and Liu, Shuo and Liu, Chao and Ye, Zichun and Ji, Jianmin and Zhao, Jie and Zhang, Yu and others},
  booktitle={18th USENIX Symposium on Operating Systems Design and Implementation (OSDI 24)},
  pages={289--305},
  year={2024}
}

@inproceedings{simCLR,
  title={A simple framework for contrastive learning of visual representations},
  author={Chen, Ting and Kornblith, Simon and Norouzi, Mohammad and Hinton, Geoffrey},
  booktitle={International conference on machine learning},
  pages={1597--1607},
  year={2020},
  organization={PmLR}
}

@inproceedings{feng2020codebert,
  title={Codebert: A pre-trained model for programming and natural languages},
  author={Feng, Zhangyin and Guo, Daya and Tang, Duyu and Duan, Nan and Feng, Xiaocheng and Gong, Ming and Shou, Linjun and Qin, Bing and Liu, Ting and Jiang, Daxin and others},
  booktitle={Findings of the association for computational linguistics: EMNLP 2020},
  pages={1536--1547},
  year={2020}
}

@inproceedings{TransH,
  title={Knowledge graph embedding by translating on hyperplanes},
  author={Wang, Zhen and Zhang, Jianwen and Feng, Jianlin and Chen, Zheng},
  booktitle={Proceedings of the AAAI conference on artificial intelligence},
  volume={28},
  number={1},
  year={2014}
}

@article{LambdaRank,
  title={Learning to rank with nonsmooth cost functions},
  author={Burges, Christopher and Ragno, Robert and Le, Quoc},
  journal={Advances in neural information processing systems},
  volume={19},
  year={2006}
}

@manual{tensorrt-docs,
  title        = {NVIDIA TensorRT Documentation},
  author       = {{NVIDIA}},
  organization = {NVIDIA},
  year         = {2026},
  note         = {Accessed: 2026-04-12},
  url          = {https://docs.nvidia.com/deeplearning/tensorrt/latest/}
}

@manual{onednn-docs,
  title        = {oneDNN Documentation},
  author       = {{Intel}},
  organization = {Intel},
  year         = {2026},
  note         = {Accessed: 2026-04-12},
  url          = {https://www.intel.com/content/www/us/en/developer/tools/oneapi/onednn-documentation.html}
}

@manual{onnx-docs,
  title        = {ONNX Documentation},
  author       = {{ONNX Community}},
  organization = {ONNX},
  year         = {2026},
  note         = {Version 1.22.0, Accessed: 2026-04-12},
  url          = {https://onnx.ai/onnx/}
}

@inproceedings{tillet2019triton,
  title={Triton: an intermediate language and compiler for tiled neural network computations},
  author={Tillet, Philippe and Kung, Hsiang-Tsung and Cox, David},
  booktitle={Proceedings of the 3rd ACM SIGPLAN International Workshop on Machine Learning and Programming Languages},
  pages={10--19},
  year={2019}
}

@article{transE,
  title={Translating embeddings for modeling multi-relational data},
  author={Bordes, Antoine and Usunier, Nicolas and Garcia-Duran, Alberto and Weston, Jason and Yakhnenko, Oksana},
  journal={Advances in neural information processing systems},
  volume={26},
  year={2013}
}

@inproceedings{TransR,
  title={Learning entity and relation embeddings for knowledge graph completion},
  author={Lin, Yankai and Liu, Zhiyuan and Sun, Maosong and Liu, Yang and Zhu, Xuan},
  booktitle={Proceedings of the AAAI conference on artificial intelligence},
  volume={29},
  number={1},
  year={2015}
}

@inproceedings{TransD,
  title={Knowledge graph embedding via dynamic mapping matrix},
  author={Ji, Guoliang and He, Shizhu and Xu, Liheng and Liu, Kang and Zhao, Jun},
  booktitle={Proceedings of the 53rd annual meeting of the association for computational linguistics and the 7th international joint conference on natural language processing (volume 1: Long papers)},
  pages={687--696},
  year={2015}
}

@article{lattner2020mlir,
  title={MLIR: A compiler infrastructure for the end of Moore's law},
  author={Lattner, Chris and Amini, Mehdi and Bondhugula, Uday and Cohen, Albert and Davis, Andy and Pienaar, Jacques and Riddle, River and Shpeisman, Tatiana and Vasilache, Nicolas and Zinenko, Oleksandr},
  journal={arXiv preprint arXiv:2002.11054},
  year={2020}
}

@manual{cutlass-docs,
  title        = {{CUTLASS}: CUDA Templates for Linear Algebra Subroutines and Solvers},
  author       = {{NVIDIA}},
  organization = {NVIDIA},
  year         = {2026},
  note         = {Accessed: 2026-04-12},
  url          = {https://docs.nvidia.com/cutlass/}
}

@inproceedings{qiao2025pruner,
  title={Pruner: A draft-then-verify exploration mechanism to accelerate tensor program tuning},
  author={Qiao, Liang and Shi, Jun and Hao, Xiaoyu and Fang, Xi and Zhang, Sen and Zhao, Minfan and Zhu, Ziqi and Chen, Junshi and An, Hong and Tang, Xulong and others},
  booktitle={Proceedings of the 30th ACM International Conference on Architectural Support for Programming Languages and Operating Systems, Volume 2},
  pages={949--965},
  year={2025}
}

@inproceedings{jeong2025bayesian,
  title={Bayesian code diffusion for efficient automatic deep learning program optimization},
  author={Jeong, Isu and Lee, Seulki},
  booktitle={19th USENIX Symposium on Operating Systems Design and Implementation (OSDI 25)},
  pages={295--311},
  year={2025}
}

@inproceedings{ma2025intelligen,
  title={IntelliGen: Instruction-Level Auto-tuning for Tensor Program with Monotonic Memory Optimization},
  author={Ma, Zixuan and Wang, Haojie and Xing, Jingze and Huang, Shuhong and Zheng, Liyan and Zhang, Chen and Cao, Huanqi and Huang, Kezhao and Zhai, Mingshu and Tang, Shizhi and others},
  booktitle={Proceedings of the 23rd ACM/IEEE International Symposium on Code Generation and Optimization},
  pages={107--122},
  year={2025}
}

@inproceedings{lucke2025mlirtransformdialect,
  title={The MLIR transform dialect: Your compiler is more powerful than you think},
  author={L{\"u}cke, Martin Paul and Zinenko, Oleksandr and Moses, William S and Steuwer, Michel and Cohen, Albert},
  booktitle={Proceedings of the 23rd ACM/IEEE International Symposium on Code Generation and Optimization},
  pages={241--254},
  year={2025}
}

@inproceedings{tirichine2026reinforcementmlir,
  title={A Reinforcement Learning Environment for Automatic Code Optimization in the MLIR Compiler},
  author={Tirichine, Mohammed and Ameur, Nassim and Bendib, Nazim and Aouadj, Iheb Nassim and Bouchama, Djad and Bouloudene, Rafik and Baghdadi, Riyadh},
  booktitle={2026 IEEE/ACM International Symposium on Code Generation and Optimization (CGO)},
  pages={696--710},
  year={2026},
  organization={IEEE}
}

@article{jaber2026autokernel,
  title={AutoKernel: Autonomous GPU Kernel Optimization via Iterative Agent-Driven Search},
  author={Jaber, Jaber and Jaber, Osama},
  journal={arXiv preprint arXiv:2603.21331},
  year={2026}
}

@article{qu2026two,
  title={A Two-Stage GPU Kernel Tuner Combining Semantic Refactoring and Search-Based Optimization},
  author={Qu, Qiuyi and Sui, Yicheng and Sun, Yufei and Chen, Rui and Zhang, Xiaofei and Zhang, Yuzhi and Wang, Haofeng and Lan, Ge},
  journal={arXiv preprint arXiv:2601.12698},
  year={2026}
}

@manual{llvm-docs,
  title        = {LLVM Documentation},
  author       = {{LLVM Project}},
  organization = {LLVM Project},
  year         = {2026},
  note         = {Accessed: 2026-04-12},
  url          = {https://llvm.org/docs/}
}

@manual{nvcc-docs,
  title        = {NVIDIA CUDA Compiler Driver NVCC Documentation},
  author       = {{NVIDIA}},
  organization = {NVIDIA},
  year         = {2026},
  note         = {Version 13.1, Accessed: 2026-04-12},
  url          = {https://docs.nvidia.com/cuda/cuda-compiler-driver-nvcc/index.html}
}

@article{zhang2023buddycompiler,
  title={Compiler Technologies in Deep Learning Co-Design: A Survey},
  author={Zhang, Hongbin and Xing, Mingjie and Wu, Yanjun and Zhao, Chen},
  journal={Intelligent Computing},
  year={2023},
  publisher={AAAS}
}

@article{gpt4,
  title={Gpt-4 technical report},
  author={Achiam, Josh and Adler, Steven and Agarwal, Sandhini and Ahmad, Lama and Akkaya, Ilge and Aleman, Florencia Leoni and Almeida, Diogo and Altenschmidt, Janko and Altman, Sam and Anadkat, Shyamal and others},
  journal={arXiv preprint arXiv:2303.08774},
  year={2023}
}

@misc{claude3,
  title        = {The Claude 3 Model Family: Opus, Sonnet, Haiku},
  author       = {{Anthropic}},
  year         = {2024},
  note         = {Model card},
  url          = {https://www-cdn.anthropic.com/de8ba9b01c9ab7cbabf5c33b80b7bbc618857627/ModelCardClaude3.pdf}
}

@article{qwen25,
  title   = {Qwen2.5 Technical Report},
  author  = {{Qwen} and Yang, An and Yang, Baosong and Zhang, Beichen and Hui, Binyuan and Zheng, Bo and Yu, Bowen and Li, Chengyuan and Liu, Dayiheng and Huang, Fei and Wei, Haoran and others},
  journal = {arXiv preprint arXiv:2412.15115},
  year    = {2024},
  url     = {https://arxiv.org/abs/2412.15115}
}

@article{deepseekv3,
  title={Deepseek-v3 technical report},
  author={Liu, Aixin and Feng, Bei and Xue, Bing and Wang, Bingxuan and Wu, Bochao and Lu, Chengda and Zhao, Chenggang and Deng, Chengqi and Zhang, Chenyu and Ruan, Chong and others},
  journal={arXiv preprint arXiv:2412.19437},
  year={2024}
}

@article{lecun2022path,
  title={A path towards autonomous machine intelligence version 0.9. 2, 2022-06-27},
  author={LeCun, Yann and others},
  journal={Open Review},
  volume={62},
  number={1},
  pages={1--62},
  year={2022}
}

@inproceedings{bi2026motus,
  title={Motus: A unified latent action world model},
  author={Bi, Hongzhe and Tan, Hengkai and Xie, Shenghao and Wang, Zeyuan and Huang, Shuhe and Liu, Haitian and Zhao, Ruowen and Feng, Yao and Xiang, Chendong and Rong, Yinze and others},
  booktitle={Proceedings of the IEEE/CVF Conference on Computer Vision and Pattern Recognition},
  pages={35101--35113},
  year={2026}
}

@article{taniguchi2026generative,
  title={Generative emergent communication: Large language model is a collective world model},
  author={Taniguchi, Tadahiro and Ueda, Ryo and Nakamura, Tomoaki and Suzuki, Masahiro and Taniguchi, Akira},
  journal={Advanced Robotics},
  pages={1--26},
  year={2026},
  publisher={Taylor \& Francis}
}
